\documentclass{article} % For LaTeX2e
\usepackage{main,times}
\iclrfinalcopy 

\usepackage{amsmath,amsfonts,bm}

\def\eqref#1{equation~\ref{#1}}
\def\1{\bm{1}}

\DeclareMathAlphabet{\mathsfit}{\encodingdefault}{\sfdefault}{m}{sl}
\SetMathAlphabet{\mathsfit}{bold}{\encodingdefault}{\sfdefault}{bx}{n}

\definecolor{ICLRblue}{rgb}{0.21,0.49,0.74}
\usepackage[pagebackref,colorlinks,allcolors=ICLRblue]{hyperref}
\usepackage{url}
\usepackage{graphicx}
\usepackage{subcaption}
\usepackage{tabularx}

\usepackage[utf8]{inputenc} % allow utf-8 input
\usepackage[T1]{fontenc}    % use 8-bit T1 fonts
\usepackage{booktabs,tabulary,multirow,overpic,xcolor,calc}
\usepackage{colortbl}
\usepackage{enumitem}
\usepackage[table]{xcolor}
\usepackage{makecell}
\usepackage{algorithm}
\usepackage{algpseudocode}
\usepackage{subcaption}
\usepackage{wrapfig}
\definecolor{color1}{rgb}{0.95,0.95,0.95}
\definecolor{color2}{rgb}{0.858, 0.188, 0.478}
\definecolor{color3}{rgb}{0.95,0.95,0.95}
\definecolor{rouse}{rgb}{0.981,0.961,0.941}
\definecolor{light-yellow}{rgb}{1,1,0.93}
\definecolor{light-green}{rgb}{0.95,1,0.95}

\title{A$^2$-Edit: Precise Reference-Guided Image Editing of \underline{A}rbitrary Objects and \underline{A}mbiguous Masks}

\author{
\parbox{0.95\textwidth}{
Huayu Zheng$^{1}$, \enspace
Guangzhao Li$^{1,2}$, \enspace
Baixuan Zhao$^{1}$, \enspace
Siqi Luo$^{1}$, \enspace
Hantao Jiang$^{1}$, \\[1.2mm]
Guangtao Zhai$^{1}$, \enspace
Xiaohong Liu$^{1,2,{\dagger}}$
} \\[5mm]
\parbox{0.95\textwidth}{
\textsuperscript{1}Shanghai Jiao Tong University,\enspace
\textsuperscript{2}Shanghai Innovation Institute
}
\vspace{-6mm}
}

\iclrfinalcopy % Uncomment for camera-ready version, but NOT for submission.
\begin{document}

\maketitle
\begin{figure}[H]
  \centering
  % \vspace{-30pt}
  \includegraphics[width=\linewidth]{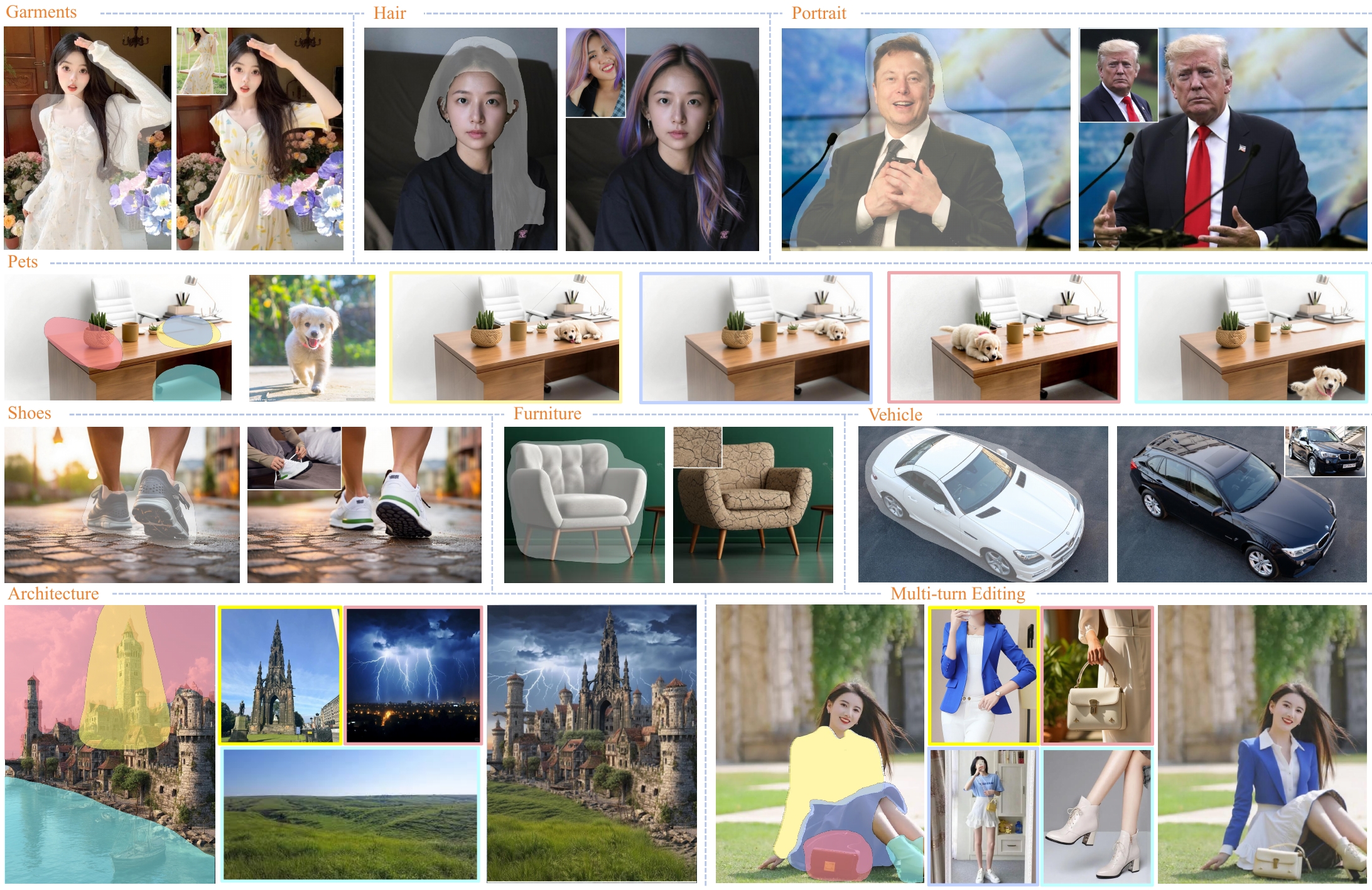}
  \vspace{-2mm}
  \caption{\textbf{Visual Results of A\texorpdfstring{$^2$}{2}-Edit}: A Unified Framework for Arbitrary-Object Inpainting. Our method supports a wide spectrum of real-world scenarios, delivering high-quality results across diverse object categories. The examples illustrate the generality and robustness of our method in handling various image editing tasks. More results are provided in the supplementary material.}
  \label{figurelabel}
\end{figure}

{
  \renewcommand{\thefootnote}%
    {\fnsymbol{footnote}}
 \footnotetext[0]{$^{\dagger}$Corresponding author.
  }
}

\begin{abstract}
We propose \textbf{A$^2$-Edit}, a unified inpainting framework for arbitrary object categories, which allows users to replace any target region with a reference object using only a coarse mask. To address the issues of severe homogenization and limited category coverage in existing datasets, we construct a large-scale, multi-category dataset \textbf{UniEdit-500K}, which includes 8 major categories, 209 fine-grained subcategories, and a total of 500,104 image pairs. Such rich category diversity poses new challenges for the model, requiring it to automatically learn semantic relationships and distinctions across categories. To this end, we introduce the \textbf{Mixture of Transformer} module, which performs differentiated modeling of various object categories through dynamic expert selection, and further enhances cross-category semantic transfer and generalization through collaboration among experts. In addition, we propose a \textbf{Mask Annealing Training Strategy} (MATS) that progressively relaxes mask precision during training, reducing the model’s reliance on accurate masks and improving robustness across diverse editing tasks. Extensive experiments on benchmarks such as VITON-HD and AnyInsertion demonstrate that A$^2$-Edit consistently outperforms existing approaches across all metrics, providing a new and efficient solution for arbitrary object editing. Code and dataset will be released upon publication.
\end{abstract}   
\section{Introduction}
\label{sec:intro}

In recent years, the rapid advancement of diffusion models~\cite{ho2020ddpm,peebles2023scalable} has significantly propelled the progress of image generation technologies. However, effectively adapting these models to precise and controllable image editing tasks~\cite{brooks2023instructpix2pix, zhang2023magicbrush, wang2024taming, shen2024audioscenic, li2024drip, xu2024gg} remains challenging. Among various editing paradigms, reference-guided image inpainting has emerged as an important direction due to its strong controllability and practical value. 
% This task requires the model to transfer semantic content from a reference image to a designated target region under mask constraints. At the same time, the model must preserve the identity of the reference object and ensure natural integration with the surrounding scene. 
This task requires transferring semantic content from a reference image to a masked target region while preserving object identity and maintaining natural consistency with the surrounding scene.
Such capability forms the core technical foundation for applications including e-commerce product placement, virtual try-on, and personalized image customization.

Despite significant progress in recent years, existing reference-guided image inpainting methods still face critical bottlenecks in practical applications, which severely limit their generalization ability and usability. On the one hand, current approaches are typically optimized for specific object domains (e.g., garments~\cite{chong2024catvton, xu2024ootdiffusion} or portraits~\cite{parihar2024text2place, kulal2023putting}), while editing tasks across different categories exhibit fundamentally different modeling objectives. For instance, garment editing emphasizes texture consistency, portrait editing focuses on identity preservation and morphological variations, and rigid objects require accurate geometric and material representation. Such disparities in objectives make it difficult for a unified model to satisfy cross-category optimization requirements through a single processing pathway. However, most existing methods still employ shared parameter pathways to process inputs from all categories, lacking category-specific modeling mechanisms, which leads to significant cross-category performance degradation or even failure. Meanwhile, existing datasets exhibit highly homogeneous category distributions and limited coverage~\cite{he2024freeedit,choi2021viton,chen2024anydoor,chen2024zero,song2025insert}, further weakening the model’s ability to learn universal representations and exacerbating the generalization bottleneck.

On the other hand, current methods generally rely on high-precision segmentation masks to strictly define editing regions~\cite{song2025insert,chen2024anydoor,chen2024zero}. In real-world scenarios, however, whether using user-drawn sketches or automatically generated detection boxes, achieving pixel-level accuracy is difficult. When mask quality deteriorates, model performance degrades significantly. Such reliance on idealized spatial guidance not only reduces interaction flexibility but also makes models fragile in open environments, thereby hindering their practical deployment and application.

To address the aforementioned challenges, we propose \textbf{A$^2$-Edit}, a unified reference-guided image inpainting framework designed for \textbf{A}rbitrary object categories and \textbf{A}rbitrary levels of mask precision. A$^2$-Edit enhances both cross-category generalization and mask robustness within a unified model architecture through dynamic expert modeling and a progressive training mechanism. Specifically, we introduce a \textbf{Mixture of Transformers (MoT)} architecture, which dynamically routes input features to specialized expert subnetworks based on the semantic characteristics and category attributes of the input object. This design enables category-specific modeling while facilitating cross-category knowledge collaboration, thereby significantly improving the generalization capability of the unified model.  Meanwhile, we propose the \textbf{Mask Annealing Training Strategy (MATS)}, which progressively reduces mask precision during training. This strategy encourages the model to shift from relying on strict geometric boundaries toward developing stronger contextual semantic understanding and structural generation capabilities, thereby improving its adaptability to coarse masks.

To support the effective training of this unified framework and mitigate the issue of homogeneous data distribution, we further construct a large-scale multi-category dataset, \textbf{UniEdit-500K}. The dataset contains over \textbf{500,000} reference–target image pairs, covering \textbf{8 major categories} and \textbf{209 fine-grained subcategories}, spanning diverse object types including texture-dominant non-rigid objects and structure-dominant rigid objects. Such diversified object distribution provides critical support for collaborative modeling under heterogeneous editing objectives and enables strong cross-category generalization within a unified model. 
% To address these challenges, we propose A$^2$-Edit, a generalized reference-guided inpainting framework designed to handle arbitrary object categories and arbitrary mask precision. A$^2$-Edit introduces a Mixture of Transformers (MoT) architecture, which dynamically routes input features to specialized expert subnetworks based on the semantic characteristics and category attributes of the input object. Through differentiated modeling and cross-category collaborative learning, the framework enhances the generalization capability of a unified model. To support effective training of this architecture and alleviate data homogeneity, we construct a large-scale multi-category dataset, UniEdit-500K, covering 8 major categories and 209 fine-grained subcategories. The dataset provides diverse cross-domain image pairs with corresponding mask annotations, establishing a solid data foundation for learning cross-category semantic transfer and structural modeling capabilities.

% Furthermore, to reduce the model’s dependence on idealized high-precision masks, we propose the Mask Annealing Training Strategy (MATS). By progressively reducing mask precision during training, the model is guided to shift from relying on strict geometric boundaries toward developing stronger contextual semantic understanding and structural generation capability. This progressive training paradigm effectively improves robustness, enabling stable editing performance under varying levels of mask quality.

In summary, our main contributions are:
\begin{itemize}
\item We propose $A^2$-Edit, a highly generalized reference-guided image inpainting framework that breaks the constraints of domain-specific editing and rigid spatial guidance.
\item We introduce a novel Mixture of Transformers (MoT) architecture and a Mask Annealing Training Strategy (MATS), enabling superior cross-category semantic transfer and robust performance with arbitrarily imprecise masks.
\item We construct UniEdit-500K, the most comprehensive dataset to date for this task, covering 8 major domains and 209 subcategories to support universal editing research. Extensive evaluations show that our approach achieves state-of-the-art performance across multiple datasets and objective metrics.

\end{itemize}

\section{Related Works}
\subsection{Image Editing}
Diffusion models have surpassed traditional generative models (VAEs, GANs)~\cite{RPSRMD,pred,fastllve,raw-vsr,griddehaze,griddehaze+} in image quality, establishing themselves as the mainstream choice for editing~\cite{a3gan,ciagan,dalle,glide,stsr}. Early U-Net-based methods~\cite{ho2020ddpm,  meng2021sdedit, avrahami2022blendeddiffusion, brooks2023instructpix2pix, xu2024ootdiffusion} often suffer from editing failures and detail distortions despite fine-grained masks. While Diffusion Transformers (DiTs)~\cite{peebles2023scalable, sd3, Flux} have improved fidelity and controllability, recent text-driven approaches (IC-Edit~\cite{zhang2025context}, DreamO~\cite{mou2025dreamo}, X2I~\cite{ma2025x2i}, Qwen-Image~\cite{wu2025qwen}) still struggle with precise detail control. Mask-guided methods like Insert Anything~\cite{song2025insert} and ACE++~\cite{mao2025ace++} achieve strong performance on rigid objects, yet remain inadequate for non-rigid subjects (e.g., human faces, pets) due to complex deformations and identity sensitivity. We propose A$^2$-Edit, a reference-based local editing framework that employs multi-precision mask fine-tuning to enable high-quality, fine-grained editing of diverse objects at arbitrary regions.
\subsection{Universal Image Editing}
Recent works have explored unified editing frameworks: ACE++~\cite{mao2025ace++} unifies inputs via concatenated feature maps; Insert Anything expands training data across humans, objects, and garments; IC-Edit simulates in-context learning textually; UniReal~\cite{chen2025unireal} models editing as discontinuous video generation. However, these methods remain predominantly confined to rigid-object editing and struggle with non-rigid subjects requiring strong identity preservation. Text-driven approaches~\cite{ma2025x2i, mou2025dreamo} fail to maintain subject identity, while mask-guided methods~\cite{chen2024anydoor, chen2024zero, mao2025ace++, parihar2024text2place,he2024freeedit} are constrained by mask precision. We propose A$^2$-Edit with an Mixture of Transformers (MoT) architecture that dynamically activates specialized experts based on semantic content and deformation characteristics, enabling a unified model that preserves structural stability for rigid objects and identity consistency for non-rigid subjects.

\section{Method}
% \subsection{Preliminary}

\subsection{Overview}
\begin{figure*}
    \centering
    \includegraphics[width=\linewidth]{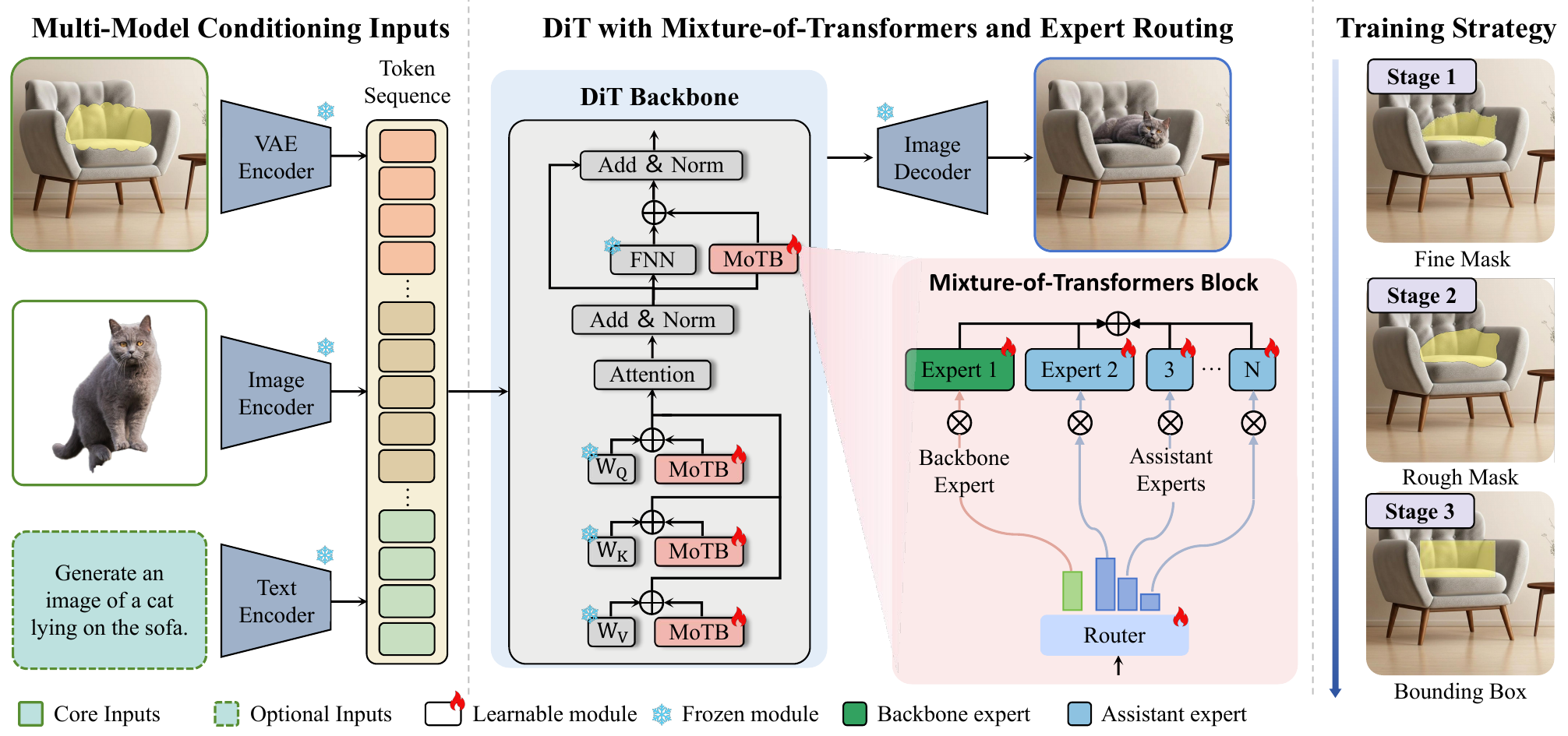}
    \caption{\textbf{Overview of the A$^2$-Edit framework.} Our architecture takes a reference image, target image, and user mask as core inputs, encodes them into a unified feature space, and feeds the fused features into Mixture-of-Transformers (\textbf{MoT}) module. The MoT Block (\textbf{MoTB}) denotes an expert-specific Transformer block embedded within the attention and feed-forward layers. The model then dynamically routes features to specialized experts for differentiated modeling. Model is trained via Mask Annealing Training Strategy (\textbf{MATS}), a three-stage training process. Final outputs are decoded by a VAE decoder for high-fidelity results.}
    \label{fig:overall}
\end{figure*}

The goal of image editing is to generate high-quality results in the target image by seamlessly integrating the subject from the reference image into the target scene, while adhering to the user-provided mask constraints and preserving the subject’s identity even when the mask is imprecise or rough. As illustrated in Figure~\ref{fig:overall}, our method integrates two key components: (1) a Mixture-of-Transformers (MoT) editing module that dynamically routes input features to different expert networks for differentiated modeling of diverse objects (\S\ref{MoT}); and (2) a Mask Annealing Training Strategy (MATS) that gradually reduces mask precision to shift the model from relying on strict spatial constraints to developing stronger contextual understanding and content generation capabilities (\S\ref{MATS}). Working jointly, these components enable the model to handle complex, multi-category image editing tasks by maintaining fine-grained details and identity features of the reference element, while producing high-quality, naturally integrated results even under relaxed mask constraints.

\subsection{Mixture of Transormers}
\label{MoT}

Existing image editing methods typically adopt a unified parameter pathway to process inputs from different object categories. However, such a design struggles to accommodate diverse editing objectives and often exhibits poor generalization to unseen categories. In recent years, the \textit{Mixture-of-Experts} (MoE) paradigm~\cite{jacobs1991adaptive,han2025vimoe,dai2024deepseekmoe} has been widely adopted in large-scale models, but sparse experts are typically introduced only in the FFN layers, while the Attention modules remain shared.

In image editing tasks, different object types require not only \textit{category-specific content transformations} (handled by FFN), but also \textit{specialized relational modeling} (handled by Attention), such as shape alignment, texture consistency, and spatial coherence. Therefore, applying expert modeling only to the FFN layers is often insufficient to capture these structural relationships. Meanwhile, recent studies~\cite{yang2025umoe, jin2410moh} further show that Attention itself contains functionally decomposable subspaces and can benefit from conditional expert selection.

% Based on these observations, we adopt a \textbf{Mixture-of-Transformers (MoT)} architecture that introduces dynamic expert routing into both the Attention and Feed-Forward Network (FFN) modules of the Transformer. 
Based on these observations, we adopt a \textbf{Mixture-of-Transformers (MoT)} architecture that extends expert routing from the conventional FFN-only design to both the Attention and Feed-Forward Network (FFN) modules.
Specifically, we introduce \textbf{Mixture-of-Transformers Blocks (MoTB)} in each Transformer layer, where multiple expert adapters are conditionally activated through a routing mechanism, enabling category-aware parameter specialization. Each expert is implemented as a lightweight \textbf{LoRA adapter}~\cite{hu2022lora}, 
enabling efficient parameter specialization with minimal computational overhead.

\subsubsection{Multimodal Feature Encoding.}
Given a reference image $I_r$, a target image $I_t$, and a mask $M$, we first extract the foreground of the reference image to obtain a clean subject representation $\tilde{I}_r$~\cite{chen2024anydoor, shin2024large}. The reference image is encoded to obtain visual features $F_r$, while the target image $I_t$ with mask guidance is encoded into latent features $F_t$. 
If a text prompt $P$ is provided, it is encoded into textual features $F_p$. 
All features are projected into a shared multimodal embedding space and fused into a unified token sequence:
\begin{equation}
\mathbf{x} = \mathrm{Fuse}(F_t, F_r, F_p).
\end{equation}
\subsubsection{Expert Routing.}
We employ a multilayer perceptron as the gating network $G(\cdot)$ to predict routing weights over $N$ experts. To balance model capacity and efficiency, we adopt an Anchor-Guided Routing (AGR) strategy $\mathcal{T}_{AGR}(\cdot)$. A backbone expert is always activated to provide universal knowledge, while assistant experts are activated only if their routing weights exceed the backbone's weight (or the highest-weighted expert otherwise). The activated expert set is defined as $\mathcal{S} = \mathcal{T}_{AGR}(\mathbf{w}(\mathbf{x}))$.

% \begin{equation}
% \mathcal{S} = \mathcal{T}_{AGR}(\mathbf{w}(\mathbf{x})).
% \end{equation}

\subsubsection{MoT Layer with LoRA Experts.}
For each linear transformation $W$ in the Transformer layer (e.g., $W_Q, W_K, W_V$ in attention and $W_1, W_2$ in FFN), we use a shared backbone weight with expert-specific LoRA increments:
\begin{equation}
W = W^{(0)} + \sum_{i \in \mathcal{S}} w_i(\mathbf{x}) \Delta W_i, \quad
\Delta W_i = A_i B_i .
\end{equation}
% \begin{equation}
% \Delta W_i = A_i B_i,
% \end{equation}
where $W^{(0)}$ denotes the shared backbone parameters and $\Delta W_i$ represents the LoRA update of the $i$-th expert.
Notably, routing is performed only once per layer, and the same routing weights are shared by both the Attention and FFN modules to ensure consistent modeling behavior.

\subsubsection{Training and Inference.}
All routing parameters and LoRA experts are optimized jointly in an end-to-end manner. During training, routing gradients automatically encourage experts to specialize in different category patterns while preserving shared knowledge. During inference, only a small subset of experts (the backbone and a few assistant experts) are activated, enabling efficient sparse computation. For out-of-distribution inputs, the gating network selects suitable expert combinations according to feature similarity, improving generalization and robustness.

\subsection{Mask Annealing Training Strategy}
\label{MATS}

% As mentioned earlier, many models overly rely on idealized inputs, which significantly limits the user's interaction freedom and causes the model's performance to deteriorate significantly when the mask quality declines. To address this issue, we propose Progressive Mask Annealing (PMA) Training Strategy. This strategy progressively reduces the precision of the input masks, guiding the model to transition from relying on fine spatial constraints to developing stronger contextual reasoning and content generation capabilities. This enables the model to make adaptive adjustments when dealing with different objects and scenarios. The entire training process is divided into three stages: \textit{Fine Mask Training}, \textit{Augmented Rough Mask Training}, and \textit{Bounding Box Training}. 

Many models overly rely on idealized inputs, limiting user interaction freedom and degrading performance with poor mask quality. We propose the Mask Annealing Training Strategy (MATS), which gradually reduces input mask precision to guide the model from relying on fine spatial constraints to stronger contextual reasoning and content generation—enabling adaptive adjustments across objects/scenarios. It includes three training stages: \textit{Fine Mask Training}, \textit{Augmented Rough Mask Training}, and \textit{Bounding Box Training}. 

% Fine masks lay the foundation for the model's basic capabilities, rough masks enhance the model's semantic control over the overall shape of objects, and bounding box training further stimulates the model's content generation and context reasoning abilities under weak supervision conditions.

\paragraph{Fine Mask Training.}
% In the initial stage of training, we use high-precision segmentation masks that are strictly aligned with the target object to indicate the editing region. Specifically, given a target image $I_t$ and its corresponding high precision mask $M_t \in \{0,1\}^{H \times W}$, along with a reference image $I_r$, this stage aims to train the model to extract identity and detail features from $I_r$ and fuse them into the masked region of the target scene $I_t$ under the precise spatial constraints of $M_t$. This stage establishes the model’s foundational ability to model local structures, textures, and semantic consistency.

In the initial training stage, we use high-precision segmentation masks strictly aligned with the target object to indicate the editing region. The model is provided with the target image $I_t$, its high-precision mask $M_t \in \{0,1\}^{H \times W}$, and reference image $I_r$.
% this stage trains the model to extract identity/detail features from $I_r$ and fuse them into $I_t$'s masked region under $M_t$'s precise spatial constraints, establishing its foundation for modeling local structures, textures, and semantic consistency.

\paragraph{Augmented Rough Mask Training.}

To reduce overreliance on mask boundaries and enhance tolerance for rough inputs, the second stage introduces mask augmentation. First, isotropic morphological dilation is applied to the fine mask $M_t$:
\begin{equation}
M_t^{dil} = M_t \oplus B_a,
\end{equation}
where $\oplus$ denotes dilation and $B_a$ is a circular structuring element of radius $a$, simulating user-drawn edge extensions. The external contour set $\mathcal{C} = \{ \mathbf{p}_i = (x_i, y_i) \}_{i=1}^N$ is then extracted from $M_t^{dil}$.
Contour points undergo spatially correlated random displacement via Perlin noise~\cite{perlin1985noises} to emulate human drawing errors:
\begin{equation}
\Delta \mathbf{p}_i = \alpha \cdot 
\begin{bmatrix}
\mathcal{N}(x_i \cdot s, y_i \cdot s) \\
\mathcal{N}((x_i + \delta) \cdot s, (y_i + \delta) \cdot s)
\end{bmatrix},
\end{equation}
where $\alpha$ controls displacement magnitude, $s$ is noise scale, and $\delta$ decouples axes. Disturbed points are:
\begin{equation}
\hat{\mathbf{p}}_i = \Pi_{\Omega}(\mathbf{p}_i + \Delta \mathbf{p}_i),
\end{equation}
with $\Pi_{\Omega}$ projecting coordinates to the image domain. The augmented rough mask $\bar{M_t}$ is derived from $\hat{\mathcal{C}} = \{ \hat{\mathbf{p}}_i \}$.

% This stage disrupts precise mask boundaries, prompting the model to shift from spatial shortcuts to semantic understanding, improving tolerance for blurred/broken/offset edges.
\paragraph{Bounding Box Training.}
% In the final stage, we further simplify the mask by extending the rough mask $\bar{M_t}$ to the bounding box of the target object $\tilde{M_t}$. At this stage, the model needs to independently infer the object’s pose, scale, and local structure in the absence of fine shape priors, and generate reasonable and naturally integrated inserted content in combination with the reference image. This phase aims to further enhance the model’s creative generation capability and strengthen its adaptability to objects like human faces and pets, where only rough localization information is provided.

In the final stage, we further simplify the mask by extending the rough mask $\bar{M_t}$ to the target object's bounding box $\tilde{M_t}$. Without fine shape priors, the model infers the object’s pose, scale, and local structure independently, generating reasonable, naturally integrated content with the reference image. This phase enhances creative generation capability and adaptability to objects (e.g., human faces, pets) with only rough localization.

% The Multi-Stage Multi-Precision Mask Training Strategy we propose is a general training paradigm for image inpainting models. This strategy employs adopts a phased progressive training mechanism to guide the model’s capabilities to improve incrementally, with training objectives gradually shifting from precise region filling toward semantically guided generation. While preserving the model’s high-fidelity detail reconstruction capability, it enhances its robustness to input masks of varying precision, thereby laying a solid foundation for uniformly handling generalized image insertion tasks involving both rigid and non-rigid objects.

\section{UniEdit-500K}
\begin{figure*}[t]
  \centering  \includegraphics[width=\textwidth]{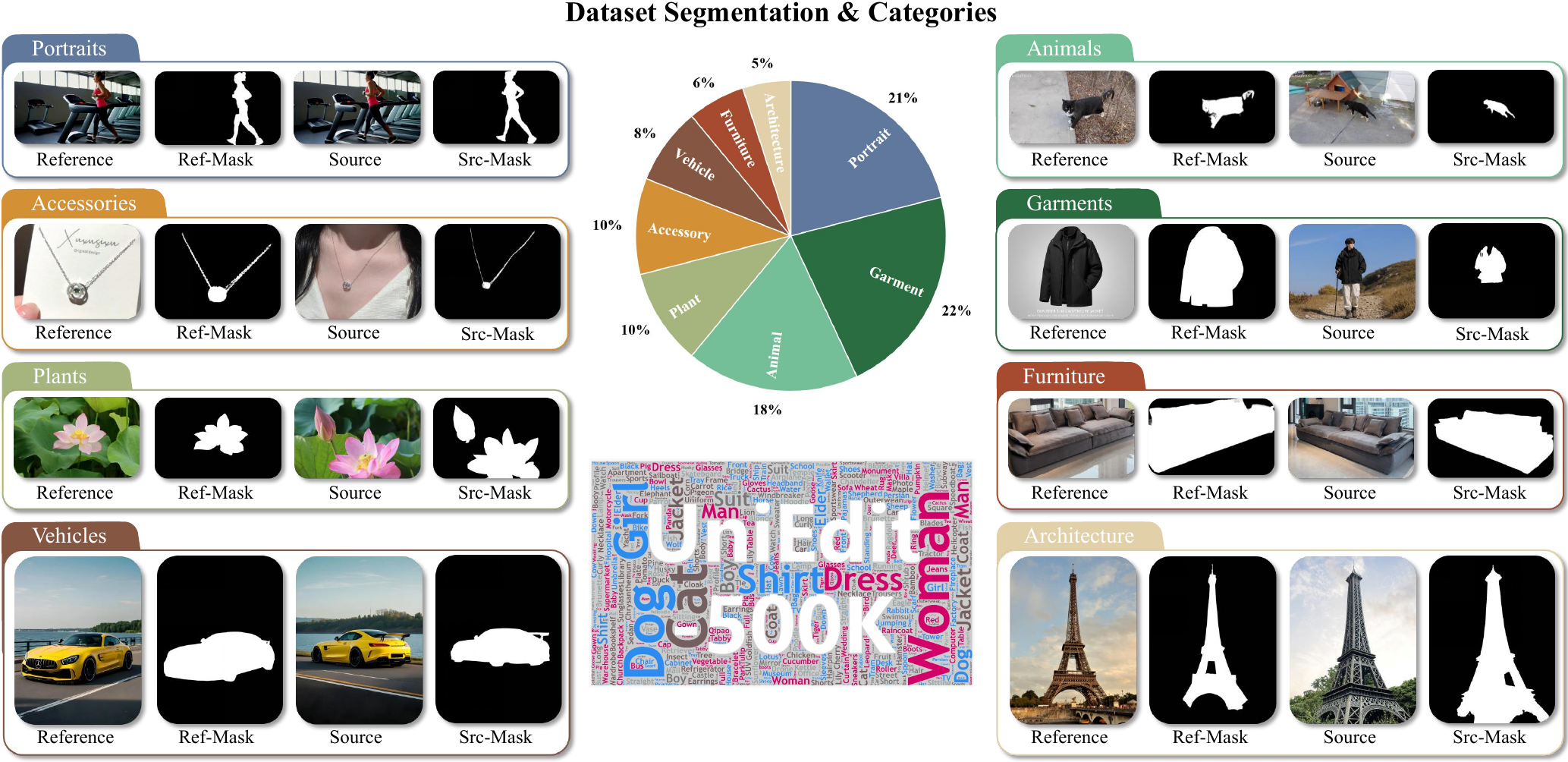}
   \caption{\textbf{An overview of our UniEdit-500K dataset.} The central pie chart illustrates the proportional distribution across the eight major categories. The word cloud below visualizes the diversity of object classes within the dataset. Surrounding the center are representative examples from each category (Portraits, Accessories, Plants, Vehicles, Animals, Garments, Furniture, and Architecture), showcasing the data format which includes a reference image, a source image, and their corresponding segmentation masks.}
   \label{fig:data}
\end{figure*}
\subsection{Comparison with Existing Datasets}

Recent works have made remarkable progress in the field of image generation. However, most existing methods focus on single tasks, perform poorly on out-of-distribution object categories, struggling to adapt to complex and diverse real-world scenarios. We attribute this issue partly to the constrained categories in existing datasets: the FreeEdit~\cite{he2024freeedit} dataset mainly focuses on animals and plants, while the VITON-HD~\cite{choi2021viton} dataset specializes in garments. Although AnyDoor~\cite{chen2024anydoor} and MimicBrush~\cite{chen2024zero} contain large scale data, they include very few samples related to human insertion. While AnyInsertion~\cite{song2025insert} expands categories to include humans, objects, and garments, it still falls short in category coverage. To address these limitations, we construct UniEdit-500K, a large-scale dataset covering multiple domains and diverse object types. This dataset comprises eight major categories—garments, portraits, animals, plants, accessories, furniture, vehicles, and architecture—further refined into 209 fine-grained subcategories. The diversity of data categories enhances the model's comprehensive capabilities across various object types. More importantly, such diversity provides the model with abundant examples to learn both distinctions and correlations among different categories, thereby establishing essential conditions for achieving genuine generalization.

% TODO 生成MASK的策略

\subsection{Data Construction}

We employed a multi-strategy hybrid approach to jointly construct the large-scale UniEdit-500K dataset.

For \textbf{garments and accessories} categories, we build upon the VITON-HD and DressCode datasets, and additionally collect a large number of \textit{person-wearing-item} scene images from open-licensed platforms such as Pixabay that allow research use and redistribution. Furthermore, we supplement the dataset with images sourced from community platforms under explicit creator permission. During data construction, we pair the standalone item images with their corresponding wearing-scene images to form high-quality image pairs. 

For \textbf{portraits and animals}, which requires strong identity consistency, we adopted a video-driven dynamic sampling method. The video data is sourced from the VidGen-1M dataset~\cite{tan2024vidgen} (Apache License 2.0). Given a video and a specified subject category, we first performed frame sampling and used Grounding DINO~\cite{liu2024grounding} and Segment Anything Model (SAM)~\cite{kirillov2023segment} to extract subject masks; then, we computed quality scores via no-reference image quality assessment to filter high-quality frames; finally, we selected the optimal two frames based on the degree of subject variation as the output.

For \textbf{plants, vehicles, architecture, and furniture} objects, we collected multi-view images from and open data platforms, combining manual verification and automatic pairing strategies to construct image pairs with sufficient viewpoint variation.

In addition, we utilized NanoBanana to address missing data and balance the dataset for certain domains. After obtaining matched image pairs, we performed object detection using Grounding DINO with the obtained object category $c$, producing bounding boxes $b_i$. Then, based on $b_i$, we generated binary segmentation masks $M_i \in \{0,1\}^{H \times W}$ using the SAM. Ultimately, we obtained paired reference and target images along with their corresponding masks, as illustrated in Figure~\ref{fig:data}. For images obtained with creator authorization, we strictly restrict usage to the explicitly permitted subset to ensure compliance and proper data governance.

% TODO: Dataset Overviewå
\subsection{Dataset Overview}
The UniEdit-500K dataset consists of a training subset and a test subset. 
The training set contains 500,104 samples covering eight major categories (clothing, portraits, animals, plants, accessories, furniture, vehicles, and buildings), comprising a total of 209 fine-grained subcategories. 
Each data sample includes a reference image, a target image, and the corresponding mask. 
The composition of the dataset is illustrated in Figure~\ref{fig:data}, with additional details provided in the supplementary materials. 
For evaluation, we constructed a test set of 200 image pairs, selecting 25 samples from each major category to comprehensively assess the model's capability across different object classes.

\section{Experiments}

\begin{figure*}[t]
  \centering
  \includegraphics[width=\textwidth]{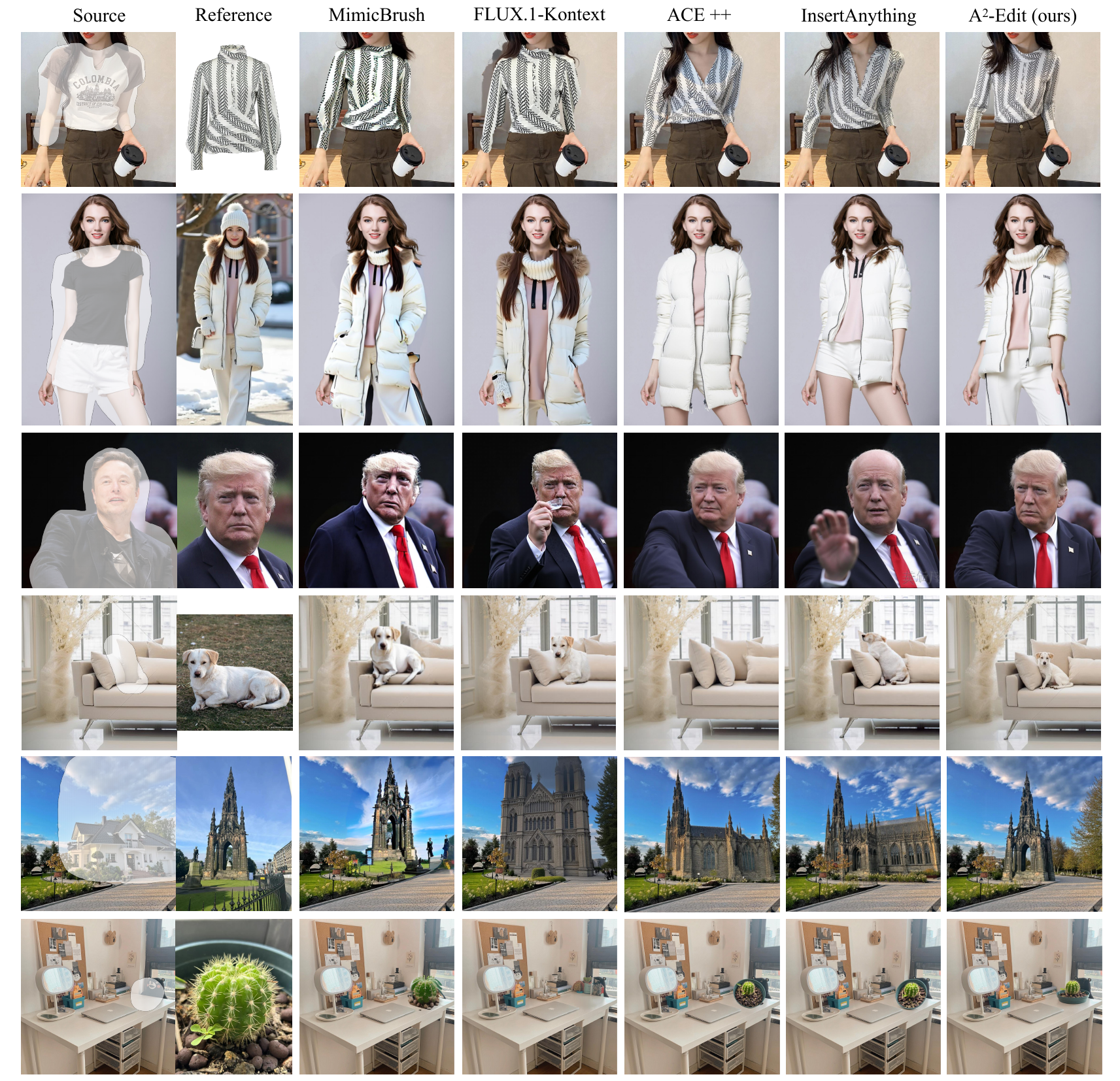}
  % \vspace{-2mm}
  \caption{\textbf{Qualitative comparison with existing mask-guided image editing methods.} Our method consistently produces more coherent structures, sharper details, and better semantic alignment with the target edit compared to existing methods( MimicBrush~\cite{chen2024zero}, FLUX.1-Kontext~\cite{flux-kontext}, ACE++~\cite{mao2025ace++} and InsertAnything~\cite{song2025insert}). }
   \label{fig:quality_1}
\end{figure*}

\subsection{Experimental Setup and Configuration}
\textbf{Implementation Details.}
We adopt FLUX.1 Fill~\cite{FluxFill} as the backbone of our approach, which is an open-source image inpainting model based on the DiT architecture. The encoder utilizes a T5 text encoder~\cite{raffel2020exploring} and a SigLIP image encoder~\cite{zhai2023sigmoid}. The model is configured with LoRA of rank 32, 8 expert subnetworks, and the Anchor-Guided Routing strategy—where one backbone expert is always activated, and assistant experts are activated based on routing weights. During training, the batch size is set to 8, and all images are processed at a resolution of 1024×1024. All experiments are conducted on a cluster of 4 NVIDIA A100 GPUs. We use our self-constructed UniEdit-500K dataset as the main training set and train the model for 6000 steps in total, including 3000 steps with fine masks, 1500 steps with augmented rough masks, and 1500 steps with bounding box masks. In terms of computational cost, during inference on a single NVIDIA A100 (80GB), our model reaches a peak memory usage of 42 GB (42,074 MiB), which is slightly higher than the 39,884 MiB required by the baseline model. For 50 sampling steps, the average inference time of our model is approximately 30 seconds, compared to about 32 seconds for the baseline model (both results are averaged over 100 evaluation samples).

\textbf{Evaluation Settings.}
We evaluate our method on two standard datasets: VITON-HD~\cite{choi2021viton} and AnyInsertion~\cite{song2025insert}. VITON-HD serves as the standard benchmark for virtual try-on and garment insertion tasks. The AnyInsertion test set contains 100 samples, including 40 object samples, 30 garment samples, and 30 portrait samples. In addition, to comprehensively validate cross-category generalization capability, we construct a test set of 200 samples by selecting 25 samples from each of the eight major categories in UniEdit-500K. For evaluation metrics, we adopt five quantitative measures: CLIP-I~\cite{clip}, DINO-I~\cite{oquab2023dinov2}, LPIPS, FID, and a VLM-based score. CLIP-I, DINO-I, and LPIPS are used to assess semantic alignment and perceptual similarity from multiple perspectives, while FID evaluates the overall distributional realism of the generated images.
Moreover, since metrics such as CLIP, DINO, and LPIPS do not always adequately capture boundary artifacts, visual realism, and local consistency, we further introduce a Vision-Language Model (VLM)-based evaluation protocol. Specifically, we employ a large language model to score the generated results along three aspects: boundary quality, realism, and local consistency. The final VLM score is obtained by averaging these three dimensions across all samples. Detailed evaluation protocols and scoring prompts are provided in the supplementary material.

\begin{table*}[t]
  \centering
  \small
  \setlength\tabcolsep{2.5pt}
  \renewcommand\arraystretch{1.0}
  \caption{\textbf{Qualitative comparison with existing image editing methods.} 
  Evaluations conducted on VTON-HD and AnyInsertion datasets show that our A$^2$-Edit model significantly outperforms existing methods. 
  The best and second-best results are demonstrated in \textbf{bold} and \underline{underlined}, respectively. 
  Note that higher DINO-I and CLIP-I scores indicate better performance, while lower LPIPS values are preferred.}
  \label{tab:comparisons_1}
  \resizebox{\linewidth}{!}{
  \begin{tabular}{l|ccc|ccc|ccc|ccc}
    \toprule
    \multirow{3}{*}{\textbf{Method}} &
    \multicolumn{6}{c|}{\textbf{VTON-HD}} &
    \multicolumn{6}{c}{\textbf{AnyInsertion}} \\
    \cmidrule(lr){2-7}\cmidrule(lr){8-13}
    & \multicolumn{3}{c|}{\emph{Fine Mask}} 
    & \multicolumn{3}{c|}{\emph{Rough Mask}} 
    & \multicolumn{3}{c|}{\emph{Fine Mask}} 
    & \multicolumn{3}{c}{\emph{Rough Mask}} \\
    \cmidrule(lr){2-4}\cmidrule(lr){5-7}\cmidrule(lr){8-10}\cmidrule(lr){11-13}
    & DINO-I  & CLIP-I  & LPIPS 
    & DINO-I  & CLIP-I  & LPIPS
    & DINO-I  & CLIP-I  & LPIPS 
    & DINO-I  & CLIP-I  & LPIPS  \\
    \midrule
    FLUX.1-Fill      & 58.59 & 72.59 & 0.1241 & 46.18 & 69.04 & 0.1509 & 50.37 & 66.06 & 0.1962 & 43.57 & 62.89 & 0.2673\\
    AnyDoor          & 59.18 & 72.54 & 0.1270 & 45.99 & 69.38 & 0.1587 & 48.15 & 65.01 & 0.2070 & 42.47 & 60.67 & 0.2854\\
    MimicBrush       & 58.99 & 73.01 & 0.1022 & 49.13 & 68.54 & 0.1479 & 50.29 & 68.88 & 0.1878 & 47.89 & 65.13 & 0.2011\\
    FLUX.1-Kontext   & 60.09 & 79.44 & 0.0800 & 56.11 & 77.91 & 0.0972 & 59.49 & 74.70 & 0.1094 & 54.15  & 74.05 & 0.1384\\    
    ACE++            & \underline{62.33} & 78.00 & 0.0767 & 58.78 & \underline{78.33} & 0.0813 & 57.69 & 74.33 & 0.1241 & 56.44 & 73.80 & 0.1327\\
    Insert Anything  & 62.15 & \underline{79.96} & \underline{0.0728} & \underline{60.87} & 78.22 & \underline{0.0809} & \underline{60.90} & \underline{76.11} & \underline{0.0978} & \underline{60.50} & \underline{75.98} & \underline{0.0992}\\
    \midrule
    \textbf{A$^2$-Edit (Ours)} & 
    \textbf{64.07} & \textbf{81.53} & \textbf{0.0677} & 
    \textbf{63.79} & \textbf{80.19} & \textbf{0.0685} & 
    \textbf{61.67} & \textbf{77.45} & \textbf{0.0909} & 
    \textbf{61.73} & \textbf{77.27} & \textbf{0.0903}\\
    \bottomrule
  \end{tabular}
  }
\end{table*}

\begin{table*}[t]
  \centering
  \small
  \setlength\tabcolsep{3pt}
  \renewcommand\arraystretch{1.05}
  \caption{\textbf{Quantitative comparison on the UniEdit test set across diverse object categories.}}
  \label{tab:comparisons_2}
  \resizebox{\linewidth}{!}{
  \begin{tabular}{l|ccccc|ccccc}
    \toprule
    \multirow{3}{*}{\textbf{Method}} &
    \multicolumn{10}{c}{\textbf{UniEdit}}\\
    \cmidrule(lr){2-11}
    & \multicolumn{5}{c|}{\emph{Fine Mask}} 
    & \multicolumn{5}{c}{\emph{Rough Mask}}  \\
    \cmidrule(lr){2-6}\cmidrule(lr){7-11}
    & DINO-I $\uparrow$ & CLIP-I $\uparrow$ & LPIPS $\downarrow$ & FID $\downarrow$ & VLM $\uparrow$ 
    & DINO-I $\uparrow$ & CLIP-I $\uparrow$ & LPIPS $\downarrow$ & FID $\downarrow$ & VLM $\uparrow$ \\
    \midrule
    FLUX.1-Fill    & 45.61 & 63.69 & 0.2519 & 93.70 & 42.7 & 40.09 & 61.58 & 0.3739 & 128.91 & 26.3 \\
    AnyDoor        & 46.76 & 64.23 & 0.2534 & 92.19 & 41.0 & 40.42 & 61.39 & 0.3623 & 120.80 & 28.3 \\
    MimicBrush     & 50.02 & 68.56 & 0.2133 & 86.23 & 47.7 & 45.07 & 64.36 & 0.2871 & 100.54 & 36.3 \\
    FLUX.1-Kontext & 53.47 & 73.99 & 0.1403 & 70.54 &
    52.3 & 48.29 & 70.56 & 0.2091 & 80.94 & 54.0 \\
    ACE++          & \underline{58.13} & 74.70 & 0.1164 & 62.30 & 76.0 & \underline{56.68} & \underline{73.66} & \underline{0.1219} & \underline{63.19} & 74.7 \\
    Insert Anything & 56.22 & \underline{74.87} & \underline{0.1121} & \underline{59.99} & \underline{78.7} & 56.13 & 73.37 & 0.1243 & 61.48 & \underline{75.0} \\
    \midrule
    \textbf{A$^2$-Edit (Ours)} & 
    \textbf{61.71} & \textbf{76.89} & \textbf{0.0898} & \textbf{57.72} & \textbf{82.3} & 
    \textbf{62.28} & \textbf{77.45} & \textbf{0.0897} & \textbf{54.62} & \textbf{79.3 }\\
    \bottomrule
  \end{tabular}
  }
\end{table*}

\subsection{Quantitative Comparison}
\textbf{Common Domain Evaluation.}
In this study, we compare A$^2$-Edit with 
FLUX.1-Fill-dev~\cite{FluxFill}, AnyDoor~\cite{chen2024anydoor},  MimicBrush~\cite{chen2024zero}, 
FLUX.1-Kontext-dev~\cite{flux-kontext}, ACE++~\cite{mao2025ace++} and InsertAnything~\cite{song2025insert}) on the VITON-HD and AnyInsertion datasets, using both fine and rough masks. The rough masks are obtained by applying isotropic morphological dilation to the fine masks. As shown in Table~\ref{tab:comparisons_1}, A$^2$-Edit achieves the best performance across all metrics,  indicating that A$^2$-Edit has the ability to perform as well as specialized models in common data domain.
Additional qualitative comparisons with representative text-guided editing models,
including Qwen-Image-Edit-2509~\cite{qwen-image-edit-2509} and FLUX.2-klein-4B~\cite{flux-klein}, are provided in the supplementary
material for cross-paradigm comparison.

\textbf{Generalization Ability Evaluation.}
To test the model's ability in different object categories, we evaluate A$^2$-Edit against other methods on the UniEdit test set. As presented in Table~\ref{tab:comparisons_2}, A$^2$-Edit significantly outperforms existing approaches on this multi-category dataset while maintaining comparable performance with those in common domains. This indicates that A$^2$-Edit exhibits more stable performance and stronger generalization capability when dealing with diverse object categories in real-world scenarios, showcasing superior universality.

\begin{wrapfigure}{r}{0.7\linewidth}
  \centering
  \includegraphics[width=\linewidth]{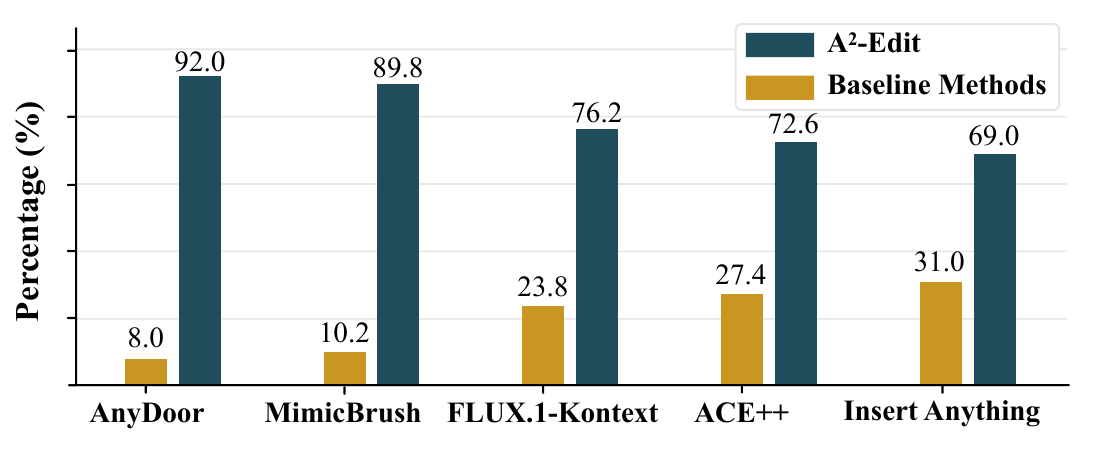}
  \caption{Statistical results of user study.}
  \vspace{-12pt}
  \label{fig:user_study}
\end{wrapfigure}
\textbf{User Study.}
In Figure~\ref{fig:user_study}, we showcase the outcomes of the results of our user study. 
A total of 24 participants were recruited to compare the images generated by our method with those produced by various baseline approaches. The reported percentages indicate the proportion of participants who preferred the results of a given model in each comparison. 
The result demonstrates that our method consistently receives higher preference from participants compared to all evaluation methods.
In Figure~\ref{fig:user_study}, we showcase the outcomes of the results of our user study. 

\subsection{Qualitative Comparison}

Figure~\ref{fig:quality_1} presents a comprehensive qualitative comparison between A$^2$-Edit and MimicBrush, FLUX.1-Kontext, ACE++, and InsertAnything across a variety of inpainting-based image editing tasks, including garments, portraits, pets, and other object categories. All experiments are conducted using hand-drawn rough masks. In several challenging editing scenarios, MimicBrush shows clear limitations in preserving fine details and fails to maintain identity consistency in portrait editing. FLUX.1-Kontext exhibits instability when handling mask boundaries, often producing noticeable edge artifacts and visible seams. InsertAnything and ACE++ frequently struggle to correctly interpret editing intentions under coarse mask conditions, generating results that remain closer to the original image style while suffering from missing details or indiscriminate filling within the masked region.

In contrast, A$^2$-Edit is able to accurately capture editing objectives across diverse object categories and complex scenes, producing semantically consistent, structurally complete, and visually stable results. For example, in garment editing tasks, it preserves the structural integrity of clothing while generating consistent texture details; in portrait replacement tasks, it maintains strong identity consistency; and in architectural replacement tasks, it effectively handles the blending between foreground and background regions. These results demonstrate the strong cross-category generalization ability of our unified editing framework. More qualitative results are provided in the supplementary material.

\subsection{Ablation Study}
\begin{figure*}[t]

  \centering
  % \captionsetup{skip=1pt}
  \includegraphics[width=\textwidth]{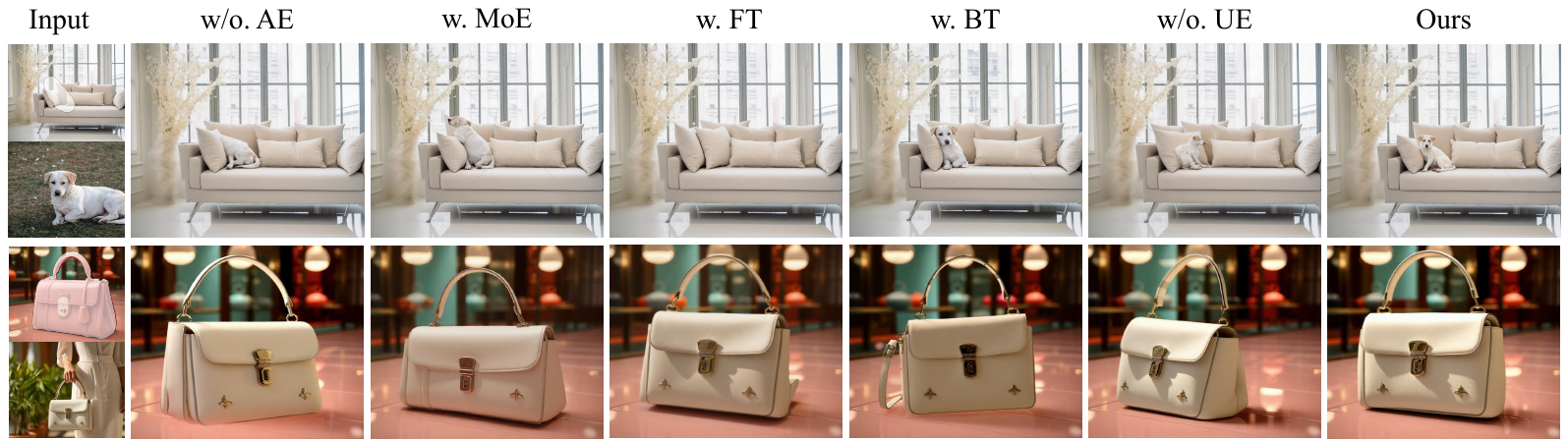}
  \caption{\textbf{A$^2$-Edit Ablation Study.} Removing the framework, training data, or MATS reduces generation quality and cross-category generalization, while progressively adding each component significantly improves detail fidelity and the model’s ability to handle rough masks.
 }
  \label{fig:ablation}
\end{figure*}

\begin{table*}[t]
  \centering
  \small
  \setlength\tabcolsep{3pt}
  \renewcommand\arraystretch{1.05}
  \caption{\textbf{Ablation Result.} We evaluate model variants on UniEdit. 
  w/o. AE: without Arbitrary Editing framework; 
  w. MoE: with MoE on FFN only;
  w/o. UE: without UniEdit-500K;
  w. FT: training with fine masks; 
  w. FT + RT: training with fine masks and augmented rough mask. 
  }
  \label{tab:ablation}
  \resizebox{\linewidth}{!}{
  \begin{tabular}{l|ccccc|ccccc}
    \toprule
    \multirow{2}{*}{\textbf{Method}} &
    \multicolumn{5}{c|}{\emph{Fine Mask}} 
    & \multicolumn{5}{c}{\emph{Rough Mask}}  \\
    \cmidrule(lr){2-6}\cmidrule(lr){7-11}
    & DINO-I $\uparrow$ & CLIP-I $\uparrow$ & LPIPS $\downarrow$ & FID $\downarrow$ & VLM $\uparrow$
    & DINO-I $\uparrow$ & CLIP-I $\uparrow$ & LPIPS $\downarrow$ & FID $\downarrow$ & VLM $\uparrow$ \\
    \midrule
    w/o. AE & 56.32 & 72.45 & 0.1423 & 75.37 & 64.3 & 56.18 & 73.21 & 0.1364 & 72.63 & 65.7 \\
    w. MoE & 56.69 & 74.01 & 0.1211 & 71.90 & 70.3 & 56.11 & 73.69 & 0.1366 & 73.04 & 70.3 \\
    w/o. UE & 58.64 & 74.32 & 0.1182 & 70.04 & 68.7 & 56.27 & 73.01 & 0.1347 & 72.84 & 66.3 \\
    w. FT  & 60.27 & 75.83 & 0.0981 & 67.55 & 71.0 & 55.65 & 72.92 & 0.1516 & 74.77 & 64.7 \\
    w. FT + RT  & 61.44 & 76.43 & 0.0941 & 63.72 & 74.7 & 59.70 & 74.86 & 0.1073 & 60.33 & 73.0 \\
    \midrule
    \textbf{Ours} & 
    \textbf{61.71} & \textbf{76.89} & \textbf{0.0898} & \textbf{57.72} & \textbf{82.3} & 
    \textbf{62.28} & \textbf{77.45} & \textbf{0.0897} & \textbf{54.62} & \textbf{79.3} \\
    \bottomrule
  \end{tabular}
  }
\end{table*}

\textbf{A$^2$-Edit framework.}
When we gradually remove the A$^2$-Edit framework and use standard MoE in NLP that only route FFNs or the baseline model for training, the test results are shown in Figure~\ref{fig:ablation}. The overall generation quality of the model decreases across all tested categories, leading to a significant drop in all quantitative metrics in Table~\ref{tab:ablation}. The standard MoE shows a noticeable improvement over the baseline model, but it still remains significantly lower than our method. This observation further validates the critical role of the A$^2$-Edit framework in achieving cross-category generalization.

\textbf{UniEdit-500K.}
As shown in Table~\ref{tab:ablation}, when our constructed UniEdit-500K training data is removed and only the untrained initial model is used for inference, all evaluation metrics exhibit a significant decline. Although the model retains a certain level of cross-category generation capability through the framework itself, it demonstrates notable deficiencies in detail generation quality and task intent comprehension. These results highlight the essential value of the UniEdit-500K dataset in enhancing the model's generalization ability and achieving cross-category object editing.

\textbf{Mask Annealing Training Strategy.}
To validate the effectiveness of MATS, we compare the model performance across four training stages: without training stage (without UniEdit-500K), training with only fine masks, training with fine masks and augmented rough masks, and the full A$^2$-Edit. As shown in Figure~\ref{fig:ablation}, when the model is trained only with fine masks, it tends to generate content excessively strictly aligned with the mask boundaries during inference with fine masks, and fails to perform meaningful edits when given rough masks due to its inability to interpret the intended editing region. After incorporating rough-mask training, this issue is significantly alleviated. Furthermore, as reported in Table~\ref{tab:ablation}, adding bbox mask training leads to additional improvements, indicating that MATS plays a crucial role in reducing the model’s dependence on precise masks and enhancing its stability and generalization across diverse object categories.

% \subsection{Comparison with Text-Guided Image Editing}
% We also make a comparison with the text-based methods. Compared with text-based control, mask-guided image editing provides clearer and more stable spatial constraints. Text-driven methods typically impose global semantic conditions on the entire image, which can easily lead to semantic leakage in complex scenes where editing a local region unintentionally alters irrelevant content. Moreover, on interaction-constrained devices such as mobile platforms, precisely describing local editing objectives and constraints through text often incurs substantial interaction costs. In contrast, masks offer a more intuitive and efficient way to specify editing regions: users can simply indicate the target area through basic gestures or rough scribbles, without requiring precise outlining or complex textual descriptions. In the supplementary material, we further compare our method with representative open-source text-guided editing models like Qwen-Image-Edit-2509~\cite{} and FLUX.2-klein-4B~\cite{}. The results show that our model consistently achieves superior performance across multiple tasks.

\section{Conclusion}
This paper presents A$^2$-Edit, a unified reference-guided image inpainting framework that overcomes the limitations of specialized approaches by supporting editing tasks across arbitrary object categories and mask precision levels. Building upon our newly constructed UniEdit-500K dataset, we implement the dynamically routed Mixture of Transformers architecture and Mask Annealing Training Strategy, effectively preserving identity consistency while enhancing the model's adaptability to diverse editing requirements and real-world input conditions. Extensive experiments across multiple benchmarks demonstrate that A²-Edit consistently achieves state-of-the-art performance across various categories, establishing a new technical standard for reference-guided image editing and providing a versatile solution for practical creative applications.

\bibliographystyle{main.bst}
\bibliography{main.bib}

@String(CVPR  = {IEEE Conf. Comput. Vis. Pattern Recog.})

@String(ICCV  = {Int. Conf. Comput. Vis.})

@String(NeurIPS = {Adv. Neural Inform. Process. Syst.})

@String(ICLR  = {Int. Conf. Learn. Represent.})

@String(ICIP  = {IEEE Int. Conf. Image Process.})

@String(ACMMM = {ACM Int. Conf. Multimedia})

@String(CVPR  = {CVPR})

@String(ICCV  = {ICCV})

@String(NeurIPS = {NeurIPS})

@String(ICLR  = {ICLR})

@String(ICIP  = {ICIP})

@String(ACMMM = {ACM MM})

@inproceedings{brooks2023instructpix2pix,
  title={Instructpix2pix: Learning to follow image editing instructions},
  author={Brooks, Tim and Holynski, Aleksander and Efros, Alexei A},
  booktitle={Proceedings of the IEEE/CVF conference on computer vision and pattern recognition},
  pages={18392--18402},
  year={2023}
}

@article{zhang2023magicbrush,
  title={Magicbrush: A manually annotated dataset for instruction-guided image editing},
  author={Zhang, Kai and Mo, Lingbo and Chen, Wenhu and Sun, Huan and Su, Yu},
  journal={Advances in Neural Information Processing Systems},
  volume={36},
  pages={31428--31449},
  year={2023}
}

@inproceedings{kulal2023putting,
  title={Putting people in their place: Affordance-aware human insertion into scenes},
  author={Kulal, Sumith and Brooks, Tim and Aiken, Alex and Wu, Jiajun and Yang, Jimei and Lu, Jingwan and Efros, Alexei A and Singh, Krishna Kumar},
  booktitle={Proceedings of the IEEE/CVF Conference on Computer Vision and Pattern Recognition},
  pages={17089--17099},
  year={2023}
}

@article{chong2024catvton,
  title={Catvton: Concatenation is all you need for virtual try-on with diffusion models},
  author={Chong, Zheng and Dong, Xiao and Li, Haoxiang and Zhang, Shiyue and Zhang, Wenqing and Zhang, Xujie and Zhao, Hanqing and Liang, Xiaodan},
  journal={arXiv preprint arXiv:2407.15886},
  year={2024}
}

@article{mao2025ace++,
  title={Ace++: Instruction-based image creation and editing via context-aware content filling},
  author={Mao, Chaojie and Zhang, Jingfeng and Pan, Yulin and Jiang, Zeyinzi and Han, Zhen and Liu, Yu and Zhou, Jingren},
  journal={arXiv preprint arXiv:2501.02487},
  year={2025}
}

@article{wang2024taming,
  title={Taming rectified flow for inversion and editing},
  author={Wang, Jiangshan and Pu, Junfu and Qi, Zhongang and Guo, Jiayi and Ma, Yue and Huang, Nisha and Chen, Yuxin and Li, Xiu and Shan, Ying},
  journal={arXiv preprint arXiv:2411.04746},
  year={2024}
}

@inproceedings{xu2024gg,
  title={GG-Editor: Locally Editing 3D Avatars with Multimodal Large Language Model Guidance},
  author={Xu, Yunqiu and Zhu, Linchao and Yang, Yi},
  booktitle={Proceedings of the 32nd ACM International Conference on Multimedia},
  pages={10910--10919},
  year={2024}
}

@article{shen2024audioscenic,
  title={AudioScenic: Audio-Driven Video Scene Editing},
  author={Shen, Kaixin and Quan, Ruijie and Zhu, Linchao and Xiao, Jun and Yang, Yi},
  journal={arXiv preprint arXiv:2404.16581},
  year={2024}
}

@article{li2024drip,
  title={Drip: Unleashing diffusion priors for joint foreground and alpha prediction in image matting},
  author={Li, Xiaodi and Yang, Zongxin and Quan, Ruijie and Yang, Yi},
  journal={Advances in Neural Information Processing Systems},
  volume={37},
  pages={79868--79888},
  year={2024}
}

@article{xu2024ootdiffusion,
  title={Ootdiffusion: Outfitting fusion based latent diffusion for controllable virtual try-on},
  author={Xu, Yuhao and Gu, Tao and Chen, Weifeng and Chen, Chengcai},
  journal={arXiv preprint arXiv:2403.01779},
  year={2024}
}

@inproceedings{choi2021viton,
  title={Viton-hd: High-resolution virtual try-on via misalignment-aware normalization},
  author={Choi, Seunghwan and Park, Sunghyun and Lee, Minsoo and Choo, Jaegul},
  booktitle={Proceedings of the IEEE/CVF conference on computer vision and pattern recognition},
  pages={14131--14140},
  year={2021}
}

@article{chen2024zero,
  title={Zero-shot image editing with reference imitation},
  author={Chen, Xi and Feng, Yutong and Chen, Mengting and Wang, Yiyang and Zhang, Shilong and Liu, Yu and Shen, Yujun and Zhao, Hengshuang},
  journal={Advances in Neural Information Processing Systems},
  volume={37},
  pages={84010--84032},
  year={2024}
}

@inproceedings{chen2024anydoor,
  title={Anydoor: Zero-shot object-level image customization},
  author={Chen, Xi and Huang, Lianghua and Liu, Yu and Shen, Yujun and Zhao, Deli and Zhao, Hengshuang},
  booktitle={Proceedings of the IEEE/CVF conference on computer vision and pattern recognition},
  pages={6593--6602},
  year={2024}
}

@article{he2024freeedit,
  title={Freeedit: Mask-free reference-based image editing with multi-modal instruction},
  author={He, Runze and Ma, Kai and Huang, Linjiang and Huang, Shaofei and Gao, Jialin and Wei, Xiaoming and Dai, Jiao and Han, Jizhong and Liu, Si},
  journal={arXiv preprint arXiv:2409.18071},
  year={2024}
}

@inproceedings{parihar2024text2place,
  title={Text2Place: Affordance-Aware Text Guided Human Placement},
  author={Parihar, Rishubh and Gupta, Harsh and VS, Sachidanand and Babu, R Venkatesh},
  booktitle={European Conference on Computer Vision},
  pages={57--77},
  year={2024},
  organization={Springer}
}

@article{song2025insert,
  title={Insert Anything: Image Insertion via In-Context Editing in DiT},
  author={Song, Wensong and Jiang, Hong and Yang, Zongxing and Quan, Ruijie and Yang, Yi},
  journal={arXiv preprint arXiv:2504.15009},
  year={2025}
}

@inproceedings{liu2024grounding,
  title={Grounding dino: Marrying dino with grounded pre-training for open-set object detection},
  author={Liu, Shilong and Zeng, Zhaoyang and Ren, Tianhe and Li, Feng and Zhang, Hao and Yang, Jie and Jiang, Qing and Li, Chunyuan and Yang, Jianwei and Su, Hang and others},
  booktitle={European Conference on Computer Vision},
  pages={38--55},
  year={2024},
  organization={Springer}
}

@inproceedings{kirillov2023segment,
  title={Segment anything},
  author={Kirillov, Alexander and Mintun, Eric and Ravi, Nikhila and Mao, Hanzi and Rolland, Chloe and Gustafson, Laura and Xiao, Tete and Whitehead, Spencer and Berg, Alexander C and Lo, Wan-Yen and others},
  booktitle={Proceedings of the IEEE/CVF international conference on computer vision},
  pages={4015--4026},
  year={2023}
}

@inproceedings{peebles2023scalable,
  title={Scalable diffusion models with transformers},
  author={Peebles, William and Xie, Saining},
  booktitle={Proceedings of the IEEE/CVF international conference on computer vision},
  pages={4195--4205},
  year={2023}
}

@inproceedings{sd3,
  title={Scaling rectified flow transformers for high-resolution image synthesis},
  author={Esser, Patrick and Kulal, Sumith and Blattmann, Andreas and Entezari, Rahim and M{\"u}ller, Jonas and Saini, Harry and Levi, Yam and Lorenz, Dominik and Sauer, Axel and Boesel, Frederic and others},
  booktitle={Forty-first international conference on machine learning},
  year={2024}
}

@misc{FluxFill,
    author={Black Forest Labs},
    title={FLUX.1-Fill-dev},
    year={2024},
    howpublished={\url{https://huggingface.co/black-forest-labs/FLUX.1-Fill-dev}},
}

@misc{Flux,
    author={Black Forest Labs},
    title={FLUX},
    year={2024},
    howpublished={\url{https://github.com/black-forest-labs/flux}},
}

@article{raffel2020exploring,
  title={Exploring the limits of transfer learning with a unified text-to-text transformer},
  author={Raffel, Colin and Shazeer, Noam and Roberts, Adam and Lee, Katherine and Narang, Sharan and Matena, Michael and Zhou, Yanqi and Li, Wei and Liu, Peter J},
  journal={Journal of machine learning research},
  volume={21},
  number={140},
  pages={1--67},
  year={2020}
}

@inproceedings{zhai2023sigmoid,
  title={Sigmoid loss for language image pre-training},
  author={Zhai, Xiaohua and Mustafa, Basil and Kolesnikov, Alexander and Beyer, Lucas},
  booktitle={Proceedings of the IEEE/CVF international conference on computer vision},
  pages={11975--11986},
  year={2023}
}

@inproceedings{perlin1985noises,
  title={An Image Synthesizer},
  author={Perlin, Ken},
  booktitle={Proceedings of the 12th Annual Conference on Computer Graphics and Interactive Techniques (SIGGRAPH)},
  pages={287--296},
  year={1985}
}

@article{shin2024large,
  title={Large-Scale Text-to-Image Model with Inpainting is a Zero-Shot Subject-Driven Image Generator},
  author={Shin, Chaehun and Choi, Jooyoung and Kim, Heeseung and Yoon, Sungroh},
  journal={arXiv preprint arXiv:2411.15466},
  year={2024}
}

@inproceedings{clip,
  title={Learning transferable visual models from natural language supervision},
  author={Radford, Alec and Kim, Jong Wook and Hallacy, Chris and Ramesh, Aditya and Goh, Gabriel and Agarwal, Sandhini and Sastry, Girish and Askell, Amanda and Mishkin, Pamela and Clark, Jack and others},
  booktitle={International conference on machine learning},
  pages={8748--8763},
  year={2021}
}

@article{oquab2023dinov2,
  title={Dinov2: Learning robust visual features without supervision},
  author={Oquab, Maxime and Darcet, Timoth{\'e}e and Moutakanni, Th{\'e}o and Vo, Huy and Szafraniec, Marc and Khalidov, Vasil and Fernandez, Pierre and Haziza, Daniel and Massa, Francisco and El-Nouby, Alaaeldin and others},
  journal={arXiv preprint arXiv:2304.07193},
  year={2023}
}

@article{zhang2025context,
  title={In-context edit: Enabling instructional image editing with in-context generation in large scale diffusion transformer},
  author={Zhang, Zechuan and Xie, Ji and Lu, Yu and Yang, Zongxin and Yang, Yi},
  journal={arXiv preprint arXiv:2504.20690},
  year={2025}
}

@article{mou2025dreamo,
  title={Dreamo: A unified framework for image customization},
  author={Mou, Chong and Wu, Yanze and Wu, Wenxu and Guo, Zinan and Zhang, Pengze and Cheng, Yufeng and Luo, Yiming and Ding, Fei and Zhang, Shiwen and Li, Xinghui and others},
  journal={arXiv preprint arXiv:2504.16915},
  year={2025}
}

@inproceedings{ma2025x2i,
  title={X2i: Seamless integration of multimodal understanding into diffusion transformer via attention distillation},
  author={Ma, Jian and Peng, Qirong and Guo, Xu and Chen, Chen and Lu, Haonan and Yang, Zhenyu},
  booktitle={Proceedings of the IEEE/CVF International Conference on Computer Vision},
  pages={16733--16744},
  year={2025}
}

@article{wu2025qwen,
  title={Qwen-image technical report},
  author={Wu, Chenfei and Li, Jiahao and Zhou, Jingren and Lin, Junyang and Gao, Kaiyuan and Yan, Kun and Yin, Sheng-ming and Bai, Shuai and Xu, Xiao and Chen, Yilei and others},
  journal={arXiv preprint arXiv:2508.02324},
  year={2025}
}

@inproceedings{chen2025unireal,
  title={Unireal: Universal image generation and editing via learning real-world dynamics},
  author={Chen, Xi and Zhang, Zhifei and Zhang, He and Zhou, Yuqian and Kim, Soo Ye and Liu, Qing and Li, Yijun and Zhang, Jianming and Zhao, Nanxuan and Wang, Yilin and others},
  booktitle={Proceedings of the Computer Vision and Pattern Recognition Conference},
  pages={12501--12511},
  year={2025}
}

@ARTICLE{RPSRMD,
  author={Wang, Wenyi and Wu, Guangyang and Cai, Weitong and Zeng, Liaoyuan and Chen, Jianwen},
  journal={IEEE Access}, 
  title={Robust Prior-Based Single Image Super Resolution Under Multiple Gaussian Degradations}, 
  year={2020},
  volume={8},
  number={},
  pages={74195-74204},
}

@INPROCEEDINGS{pred,
  author={Wu, Guangyang and Zhao, Lili and Wang, Wenyi and Zeng, Liaoyuan and Chen, Jianwen},
  booktitle=ICIP, 
  title={PRED: A Parallel Network for Handling Multiple Degradations via Single Model in Single Image Super-Resolution}, 
  year={2019}
}

@inproceedings{fastllve,
author = {Li, Wenhao and Wu, Guangyang and Wang, Wenyi and Ren, Peiran and Liu, Xiaohong},
title = {FastLLVE: Real-Time Low-Light Video Enhancement with Intensity-Aware Look-Up Table},
year = {2023},
booktitle = ACMMM,
}

@article{raw-vsr,
  author       = {Xiaohong Liu and
                  Kangdi Shi and
                  Zhe Wang and
                  Jun Chen},
  title        = {Exploit Camera Raw Data for Video Super- Resolution via Hidden Markov
                  Model Inference},
  journal      = {{IEEE} Trans. Image Process.},
  volume       = {30},
  pages        = {2127--2140},
  year         = {2021},
}

@article{griddehaze+,
  author       = {Xiaohong Liu and
                  Zhihao Shi and
                  Zijun Wu and
                  Jun Chen and
                  Guangtao Zhai},
  title        = {GridDehazeNet+: An Enhanced Multi-Scale Network With Intra-Task Knowledge
                  Transfer for Single Image Dehazing},
  journal      = {{IEEE} Trans. Intell. Transp. Syst.},
  volume       = {24},
  number       = {1},
  pages        = {870--884},
  year         = {2023},
}

@inproceedings{griddehaze,
  author       = {Xiaohong Liu and
                  Yongrui Ma and
                  Zhihao Shi and
                  Jun Chen},
  title        = {GridDehazeNet: Attention-Based Multi-Scale Network for Image Dehazing},
  booktitle    = ICCV,
  year         = {2019},
}

@article{ho2020ddpm,
  title={Denoising Diffusion Probabilistic Models},
  author={Ho, Jonathan and Jain, Ajay and Abbeel, Pieter},
  journal={NeurIPS},
  year={2020}
}

@article{hu2022lora,
  title={Lora: Low-rank adaptation of large language models.},
  author={Hu, Edward J and Shen, Yelong and Wallis, Phillip and Allen-Zhu, Zeyuan and Li, Yuanzhi and Wang, Shean and Wang, Liang and Chen, Weizhu and others},
  journal={Iclr},
  volume={1},
  number={2},
  pages={3},
  year={2022}
}

@inproceedings{meng2021sdedit,
  title={SDEdit: Guided Image Synthesis and Editing with Stochastic Differential Equations},
  author={Meng, Chenlin and Ho, Jonathan and Chen, Yang and Song, Jiaming and Ermon, Stefano},
  booktitle={ICLR},
  year={2021}
}

@article{avrahami2022blendeddiffusion,
  title={Blended Diffusion for Text-Driven Editing of Natural Images},
  author={Avrahami, Omri and Lischinski, Dani and Fried, Ohad},
  journal={CVPR},
  year={2022}
}

@inproceedings{ciagan,
  title={Ciagan: Conditional identity anonymization generative adversarial networks},
  author={Maximov, Maxim and Elezi, Ismail and Leal-Taix{\'e}, Laura},
  booktitle=CVPR,
  year={2020}
}

@inproceedings{a3gan,
  title={A3gan: {A}ttribute-aware anonymization networks for face de-identification},
  author={Zhai, Liming and Guo, Qing and Xie, Xiaofei and Ma, Lei and Wang, Yi Estelle and Liu, Yang},
  booktitle={Proceedings of the 30th ACM International Conference on Multimedia},
  pages={5303--5313},
  year={2022}
}

@article{glide,
  title={Glide: {T}owards photorealistic image generation and editing with text-guided diffusion models},
  author={Nichol, Alex and Dhariwal, Prafulla and Ramesh, Aditya and Shyam, Pranav and Mishkin, Pamela and McGrew, Bob and Sutskever, Ilya and Chen, Mark},
  journal={arXiv preprint arXiv:2112.10741},
  year={2021}
}

@article{dalle,
  title={Hierarchical text-conditional image generation with clip latents},
  author={Ramesh, Aditya and Dhariwal, Prafulla and Nichol, Alex and Chu, Casey and Chen, Mark},
  journal={arXiv preprint arXiv:2204.06125},
  volume={1},
  number={2},
  pages={3},
  year={2022}
}

@article{stsr,
  author       = {Zhihao Shi and
                  Xiaohong Liu and
                  Chengqi Li and
                  Linhui Dai and
                  Jun Chen and
                  Timothy N. Davidson and
                  Jiying Zhao},
  title        = {Learning for Unconstrained Space-Time Video Super-Resolution},
  journal      = {{IEEE} Trans. Broadcast.},
  volume       = {68},
  number       = {2},
  pages        = {345--358},
  year         = {2022},
}

@article{yang2025umoe,
  title={UMoE: Unifying Attention and FFN with Shared Experts},
  author={Yang, Yuanhang and Wang, Chaozheng and Li, Jing},
  journal={arXiv preprint arXiv:2505.07260},
  year={2025}
}

@article{han2025vimoe,
  title={Vimoe: An empirical study of designing vision mixture-of-experts},
  author={Han, Xumeng and Wei, Longhui and Dou, Zhiyang and Sun, Yingfei and Han, Zhenjun and Tian, Qi},
  journal={IEEE Transactions on Image Processing},
  volume={34},
  pages={7209--7221},
  year={2025},
  publisher={IEEE}
}

@article{dai2024deepseekmoe,
  title={Deepseekmoe: Towards ultimate expert specialization in mixture-of-experts language models},
  author={Dai, Damai and Deng, Chengqi and Zhao, Chenggang and Xu, RX and Gao, Huazuo and Chen, Deli and Li, Jiashi and Zeng, Wangding and Yu, Xingkai and Wu, Yu and others},
  journal={arXiv preprint arXiv:2401.06066},
  year={2024}
}

@article{tan2024vidgen,
  title={Vidgen-1m: A large-scale dataset for text-to-video generation},
  author={Tan, Zhiyu and Yang, Xiaomeng and Qin, Luozheng and Li, Hao},
  journal={arXiv preprint arXiv:2408.02629},
  year={2024}
}

@misc{flux-kontext,
    author={Black Forest Labs},
    title={FLUX},
    year={2025},
    howpublished={\url{https://huggingface.co/black-forest-labs/FLUX.1-Kontext-dev}},
}

@misc{qwen-image-edit-2509,
    author={Qwen},
    title={qwen},
    year={2025},
    howpublished={\url{https://github.com/QwenLM/Qwen-Image}},
}

@misc{flux-klein,
    author={Black Forest Labs},
    title={FLUX},
    year={2025},
    howpublished={\url{https://huggingface.co/black-forest-labs/FLUX.2-klein-4B}},
}

@article{jin2410moh,
  title={Moh: Multi-head attention as mixture-of-head attention, 2025},
  author={Jin, Peng and Zhu, Bo and Yuan, Li and Yan, Shuicheng},
  journal={URL https://arxiv. org/abs/2410.11842},
  year={2025}
}

@article{jacobs1991adaptive,
  title={Adaptive mixtures of local experts},
  author={Jacobs, Robert A and Jordan, Michael I and Nowlan, Steven J and Hinton, Geoffrey E},
  journal={Neural computation},
  volume={3},
  number={1},
  pages={79--87},
  year={1991},
  publisher={MIT Press}
}

\newpage
\appendix
\onecolumn
\clearpage
\setcounter{page}{1}
\appendix

\title{Supplementary for A$^2$-Edit: Precise Reference-Guided Image Editing of \underline{A}rbitrary Objects and \underline{A}mbiguous Masks} 

\maketitle

This appendix provides additional details of the UniEdit-500K dataset, including its full category breakdown and fine-grained composition. We further present extended qualitative results covering virtual try-on, identity-preserving editing, mask-guided pose control, multi-category inpainting, and multi-step editing, offering a more comprehensive evaluation of A$^2$-Edit. Additionally, we briefly compare A$^2$-Edit with text-guided image editing methods, highlighting the advantages of mask-guided approaches in terms of precise control and stability. We also introduce VLM-based evaluation metrics, further assessing the generated results in terms of boundary quality, visual realism, and local consistency. Finally, we conduct cross-category generalization evaluations, demonstrating A$^2$-Edit's strong adaptability when handling multiple categories and unseen scenarios, and discuss the method’s limitations and broader societal impact, providing a more complete picture of its applicability and societal considerations.

\section{More Details of UniEdit-500K}
The UniEdit-500K dataset contains a total of 500,104 images spanning eight major categories and 209 fine-grained subcategories. Its rich diversity and broad semantic coverage make it well-suited for training unified editing models across multiple object categories.

Among them, the \textbf{Portraits} category contains 104,706 samples, covering 22 subcategories, including Man, Woman, Boy, Girl, Baby, Elder, Blonde, Brunette, Black Hair, Red Hair, Curly Hair, Straight Hair, Long Hair, Short Hair, Front, Profile, Bust, Full Body, Sitting, Standing, Running, and Jumping.

\textbf{Garments} contains 110,508 samples, covering 26 subcategories, including T-Shirt, Shirt, Sweater, Hoodie, Jacket, Coat, Down Jacket, Suit, Outerwear, Jeans, Shorts, Trousers, Skirt, Dress, Cheongsam, Sportswear, Pajamas, Swimsuit, Vest, Cloak, Uniform, Suit, Wedding Dress, Gown, Raincoat, Windbreaker.

\textbf{Animals} contains 90,203 samples, covering 33 subcategories, including Cat, Dog, Tabby, Ragdoll, Persian, Retriever, Shepherd, Poodle, Husky, Goldfish, Hamster, Rabbit, Horse, Cow, Sheep, Pig, Tiger, Lion, Leopard, Bear, Deer, Elephant, Panda, Wolf, Bird, Eagle, Pigeon, Parrot, Duck, Goose, Chicken, Fish, Insect.

\textbf{Plants} contains 49,811 samples, covering 27 subcategories, including Tree, Shrub, Lawn, Bamboo, Pine, Cherry, Maple, Cactus, Flower, Rose, Lily, Sunflower, Tulip, Peony, Chrysanthemum, Lotus, Water Lily, Fruit, Rice, Wheat, Corn, Vegetable, Pumpkin, Tomato, Carrot, Cucumber, Bowl Picking.

\textbf{Accessories} contains 50,500 samples, covering 26 subcategories, including Hat, Glasses, Sunglasses, Earrings, Necklace, Bracelet, Ring, Hairpin, Headband, Scarf, Tie, Cap, Belt, Watch, Bag, Backpack, Wallet, Shoes, Sneakers, Heels, Boots, Mask, Gloves, Sleeves, Helmet, Umbrella.

\textbf{Furniture} contains 29,914 samples, covering 30 subcategories, including Sofa, Chair, Bed, Table, Tea Table, Desk, Bookshelf, Cabinet, Wardrobe, Rug, Lamp, Chandelier, Lamp, Mirror, Curtain, Photo Frame, Vase, Plate, Bowl, Cup, Knife, Fork, Spoon, Tray, Kettle, TV, Refrigerator, Fireplace, Washer, Computer.

\textbf{Vehicles} contains 40,107 samples, covering 24 subcategories, including Car, Sedan, SUV, Sports Car, Truck, Bus, School Bus, Motorcycle, Electromobile, Bicycle, Train, Subway, Tram, Airplane, Helicopter, Drone, Ship, Yacht, Sailboat, Speedboat, Scooter, Skateboard, Roller Blades, Tractor.

\textbf{Architecture} contains 24,355 samples, covering 21 subcategories, including House, Villa, Apartment, Office, Mall, Supermarket, Factory, Warehouse, School, Hospital, Library, Museum, Church, Temple, Bridge, Tower, Castle, Monument, Street, Square, Park.

\section{More Results}

\begin{figure*}[t]
  \centering
  \includegraphics[width=\textwidth]{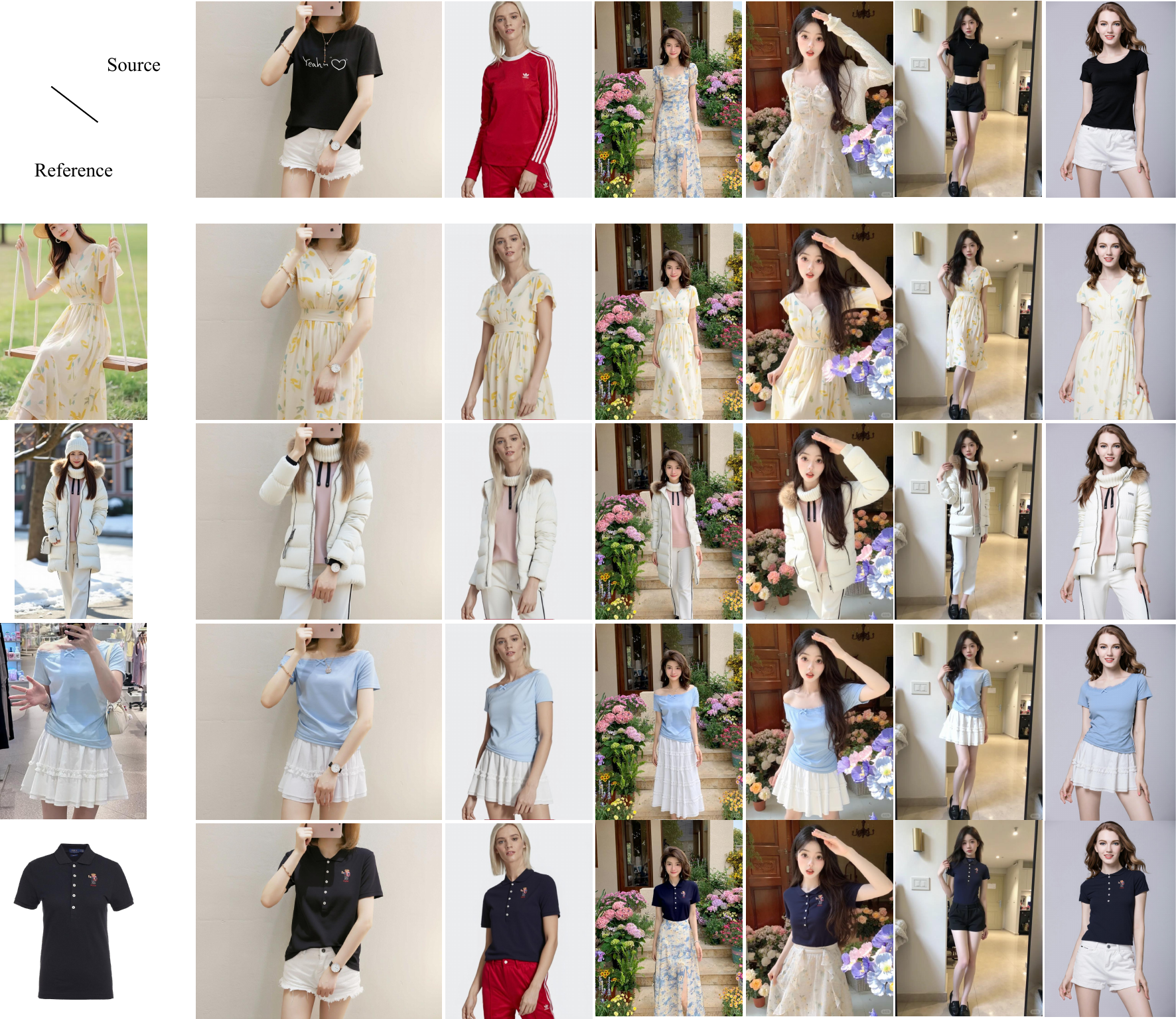}
  % \vspace{-2mm}
  \caption{Results on the virtual try-on task. The first row shows the source images, the first column shows the reference images, and the remaining images are generated by our model.}
   \label{fig:supply_1}
   \vspace{-2mm}
\end{figure*}

\begin{figure*}[t]
  \centering
  \includegraphics[width=\textwidth]{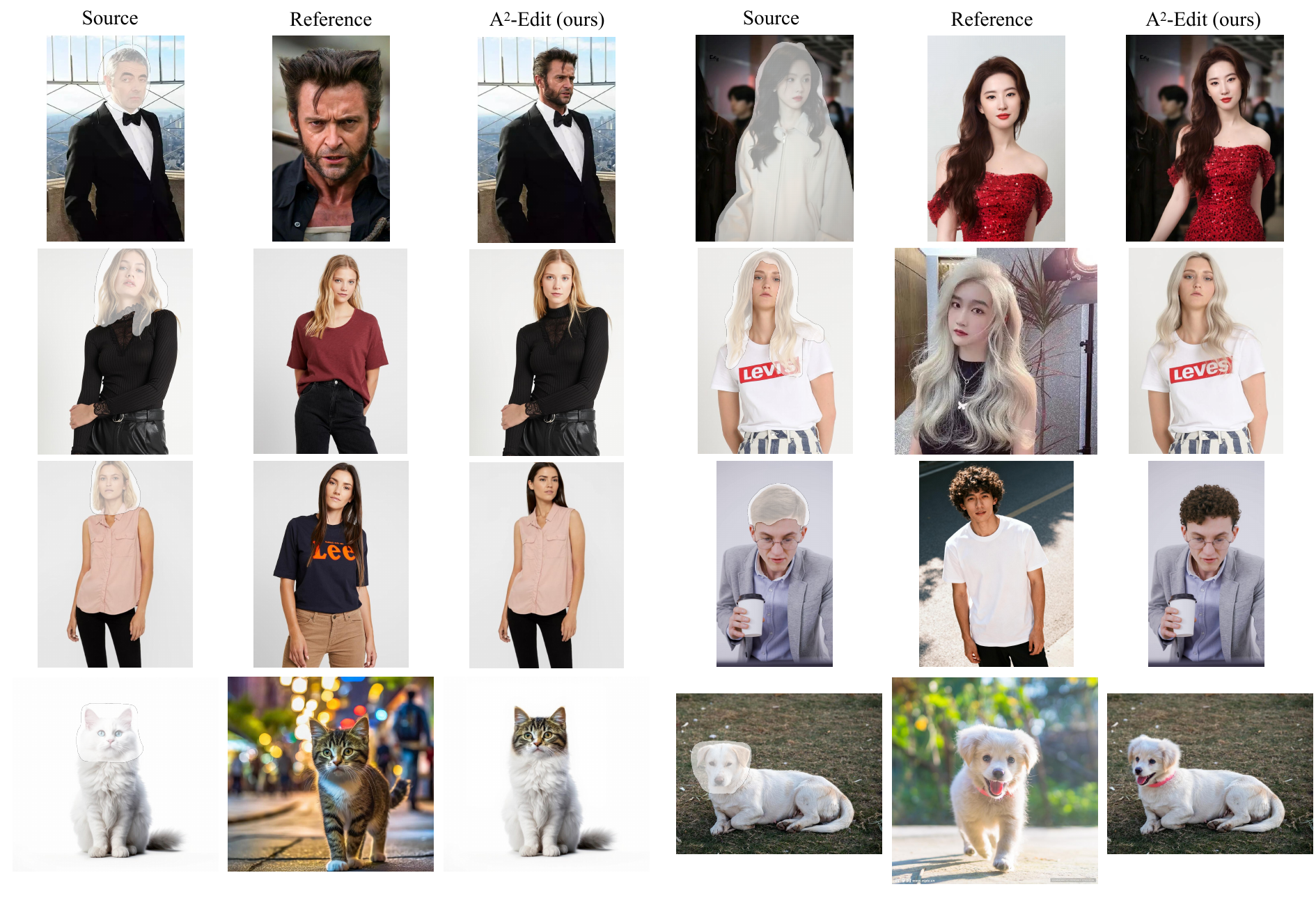}
  % \vspace{-2mm}
  \caption{Results on portraits and animals. The first three rows show human edits—including face swapping, portrait replacement, and hairstyle or color changes—where the model preserves stable identity features. The fourth row demonstrates similarly consistent results in animal face-swapping scenarios.}
   \label{fig:supply_2}
   \vspace{-2mm}
\end{figure*}

\begin{figure*}[t]
  \centering
  \includegraphics[width=\textwidth]{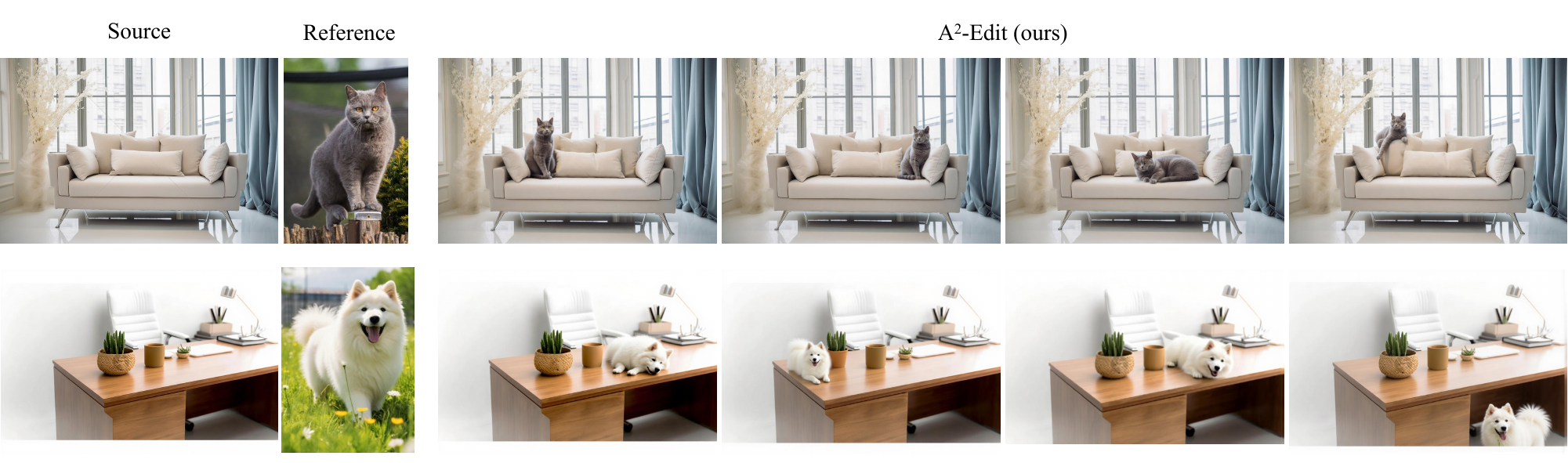}
  % \vspace{-2mm}
  \caption{Pet Results. Given different masks under a shared background, the model generates diverse poses of the same reference animal while accurately controlling position and shape.}
   \label{fig:supply_4}
   \vspace{-2mm}
\end{figure*}

\begin{figure*}[t]
  \centering
  \includegraphics[width=\textwidth]{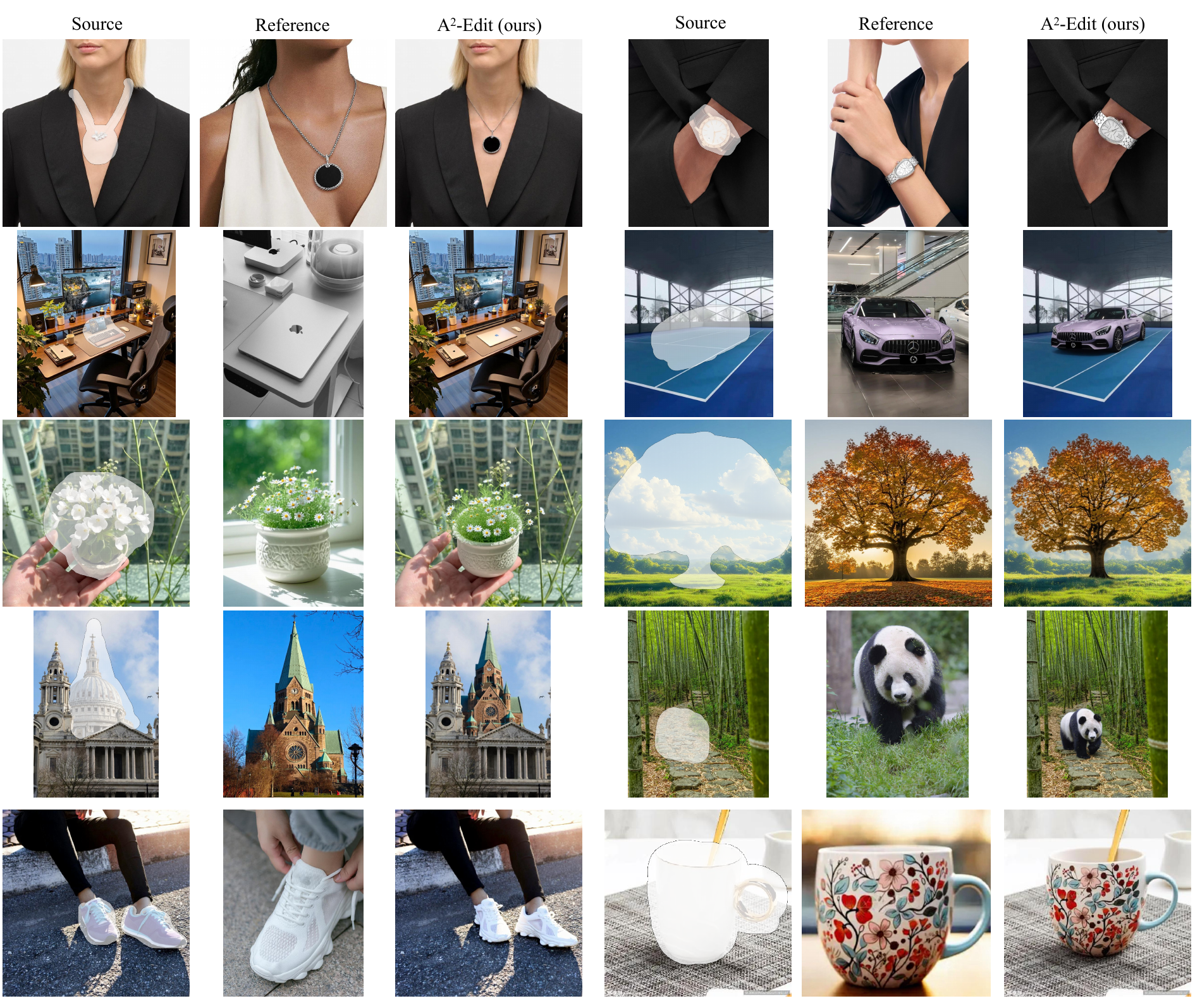}
  % \vspace{-2mm}
  \caption{More results on inpainting tasks across different object categories.}
   \label{fig:supply_3}
   \vspace{-2mm}
\end{figure*}

\begin{figure*}[t]
  \centering
  \includegraphics[width=\textwidth]{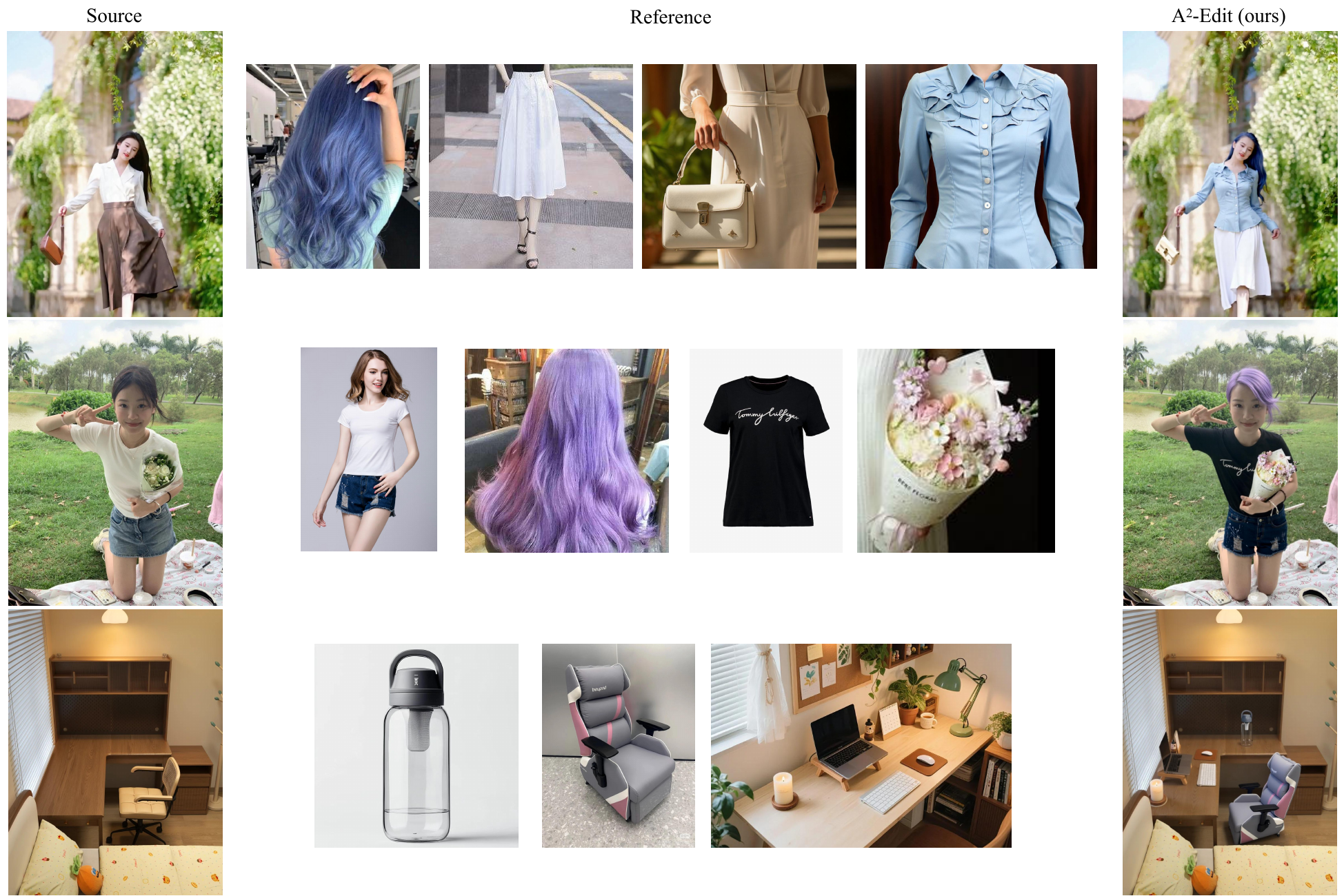}
  % \vspace{-2mm}
  \caption{Multi-step editing results. Each row represents one editing sequence, where the images in the middle correspond to the reference inputs provided at each editing step.}
   \label{fig:supply_5}
   \vspace{-2mm}
\end{figure*}

To comprehensively validate the effectiveness and generalization ability of our method, we present a wider range of diverse experimental results.

Figure~\ref{fig:supply_1} shows the model’s performance on the virtual try-on task. The first row shows the source images, the first column shows the reference images, and the remaining images are generated by our model. The results demonstrate that our method achieves stable and high-quality clothing transfer across various categories, highlighting its strong capability in garment editing tasks.

Figure~\ref{fig:supply_2} evaluates the identity consistency of our model. The first three rows focus on human subjects and examine several challenging cases, including face swapping, full portrait replacement, hairstyle modification, and hair color change. The results show that even under significant appearance alterations, the model preserves highly consistent identity features. The fourth row extends the evaluation to animals, and the results indicate that our approach also produces stable and natural results in the animal face-swapping scenario.

Figure~\ref{fig:supply_4} shows the model’s ability to generate different poses of the same reference animal under a shared background, guided by different masks. The results indicate that the model can precisely control the object’s position and shape based on the provided mask while maintaining strong identity consistency across various poses, demonstrating its advantages in spatial control and semantic coherence.

Figure~\ref{fig:supply_3} presents the model’s performance on inpainting tasks across different object categories. In all tested categories, our model produces visually coherent and semantically reasonable inpainting results.

Figure~\ref{fig:supply_5} showcases the stability of our model in multi-step editing scenarios. Even after several consecutive editing operations, the model continues to maintain high-quality and consistent results.

\begin{figure*}[t]
  \centering
  \includegraphics[width=\textwidth]{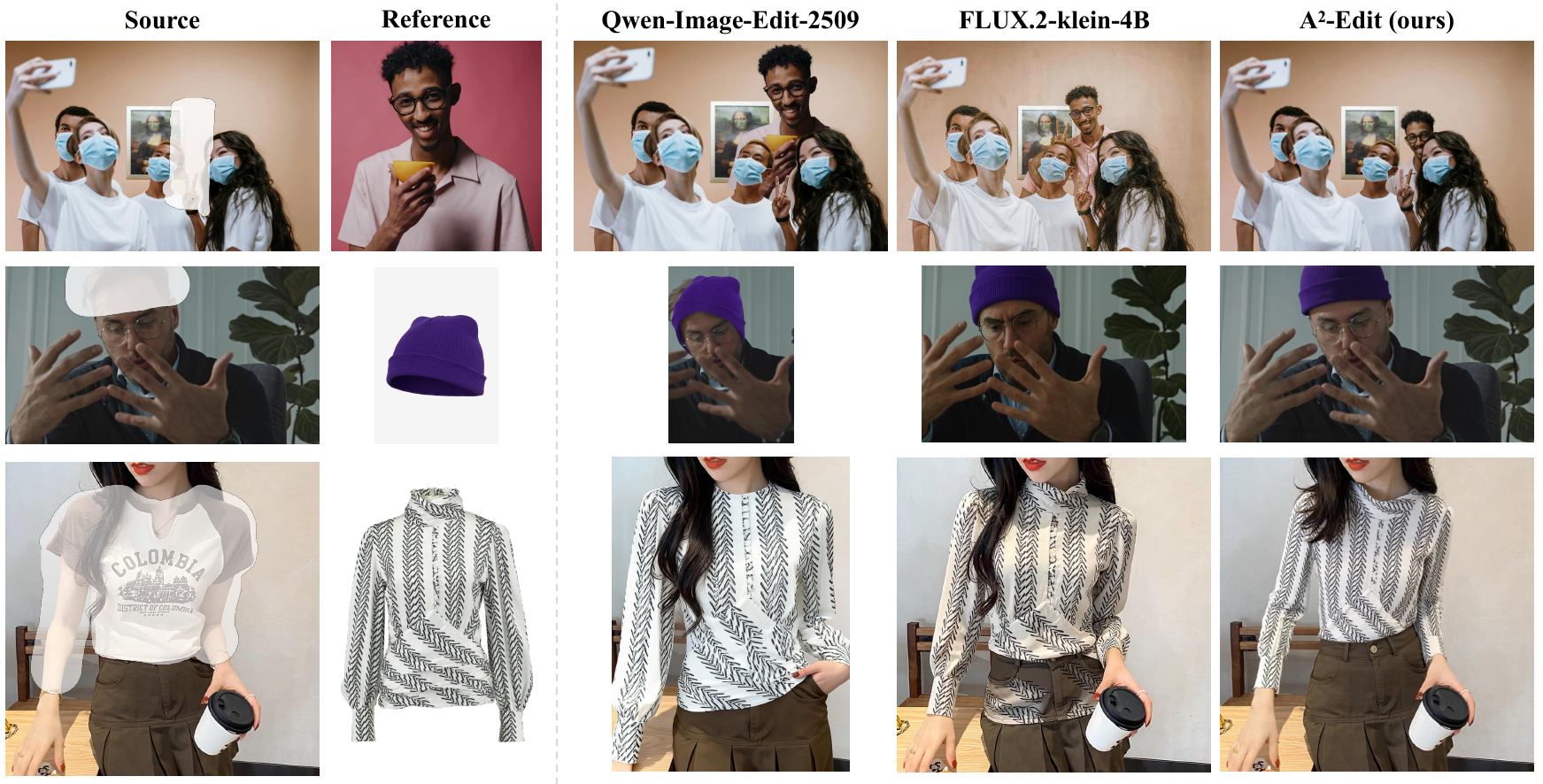}
  % \vspace{-2mm}
  \caption{Qualitative comparison with text-guided image editing models.}
   \label{fig:comparison}
   \vspace{-2mm}
\end{figure*}

\section{Comparison with Text-Guided Image Editing}
In image editing tasks, compared to relying on text prompts for control, mask-guided generation offers clearer and more stable advantages. First, text-driven methods typically impose global semantic conditions on the entire image, making it difficult to precisely constrain the editing region. Even when users intend to modify only a local area, the model may inadvertently alter the background or other irrelevant regions, thereby affecting overall consistency or disrupting fine-grained details. This phenomenon, often referred to as “semantic leakage,” becomes particularly pronounced in complex scenes.

Second, on interaction-constrained devices such as mobile platforms, precisely describing the editing region, target, and constraints through text entails considerable cognitive and input costs. Especially when fine-grained local control is required, users must provide lengthy and highly specific textual instructions, leading to a cumbersome and inefficient interaction experience.

In contrast, masks provide a more intuitive and efficient form of spatial constraint. Users can specify the editing area through simple gestures or rough scribbles, without the need for precise outlining or complex descriptions. However, in practical scenarios, requiring users to provide highly accurate masks is both fussy and unrealistic. Therefore, a generation mechanism that can effectively operate with coarse masks is of greater practical significance. In real-world applications—particularly in mobile interactive settings—coarse mask-guided generation enhances usability and user experience while maintaining strong controllability over the editing process.

Figure~\ref{fig:comparison} compares our method with recent open-source text-guided image editing models. We select Qwen-Image-Edit-2509~\cite{qwen-image-edit-2509} and FLUX.2-klein-4B~\cite{flux-klein} as representative baselines. Our model uses mask guidance, while the open-source model uses precise text prompts for control, such as "Add the purple hat from <image2> to the man's head in <image1>." In the first-row example, text-based methods exhibit unnatural blending and detail errors, such as generating three hands. In the second row, hat distortion and unintended modifications to the person’s face can be observed. In the third row, text-guided approaches fail to preserve fine-grained texture details and also introduce unnecessary changes to the subject’s pose. Overall, compared with text-guided baselines, our mask-guided framework demonstrates superior locality control and structural consistency.

\begin{figure*}[t]
  \centering
  \includegraphics[width=\textwidth]{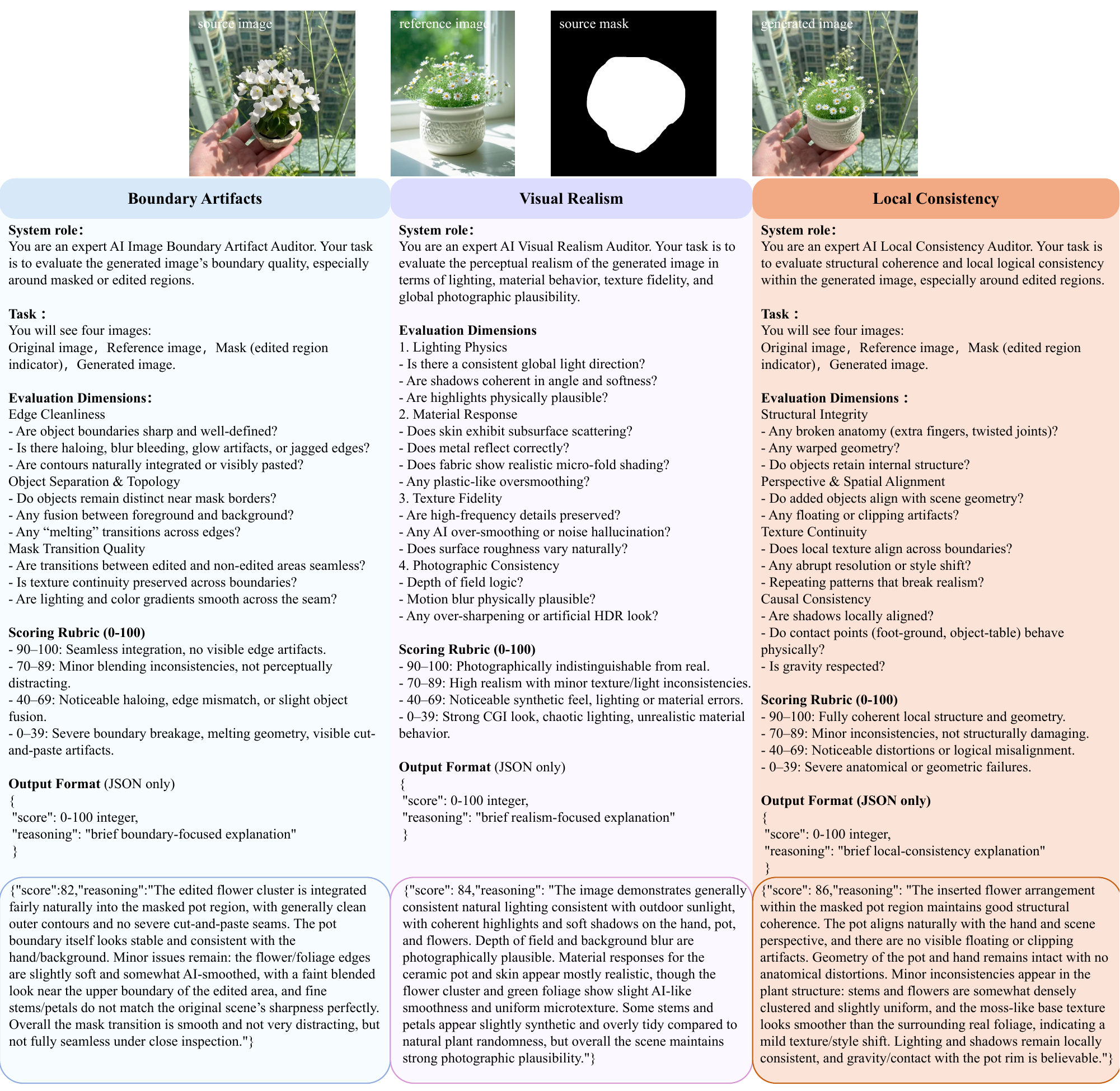}
  % \vspace{-2mm}
  \caption{VLM-based evaluation metric test example.}
   \label{fig:vlm}
   \vspace{-2mm}
\end{figure*}

\section{VLM-based Evaluation Metric Details}
Although widely used quantitative metrics such as CLIP-I, DINO-I, LPIPS, and FID measure semantic similarity, perceptual similarity, and distributional realism, they often fail to adequately capture boundary artifacts, local structural consistency, and overall visual realism in image editing tasks. Therefore, we introduce an additional Vision-Language Model (VLM)-based evaluation protocol to provide a more comprehensive assessment of the generated results.

\subsection{Evaluation Model}

We adopt a large-scale Vision-Language Model (VLM) Gemini-3 as an automated judge to evaluate the quality of generated images. The VLM receives the source image, reference image, source mask and generated image as input and assesses the editing quality based on predefined criteria. This evaluation framework enables semantic-level reasoning about image editing quality that is difficult to capture with traditional similarity metrics.

\subsection{Evaluation Dimensions}

Specifically, the VLM evaluates each generated result from three complementary aspects:

\paragraph{Boundary Quality:}
Measures whether the inserted or edited object integrates naturally with the surrounding background. This dimension evaluates the presence of artifacts, unnatural edges, or visible seams near the edited region.

\paragraph{Visual Realism}
Assesses the overall photorealism of the generated image, including texture quality, lighting consistency, and global coherence.

\paragraph{Local Consistency}
Evaluates whether the inserted object maintains structural correctness and semantic compatibility with the scene, including pose alignment, scale consistency, and object completeness.

Each dimension is scored on a scale from 1 to 100, where higher scores indicate better quality. Figure~\ref{fig:vlm} presents a test example.

\begin{figure*}[t]
  \centering
  \includegraphics[width=\textwidth]{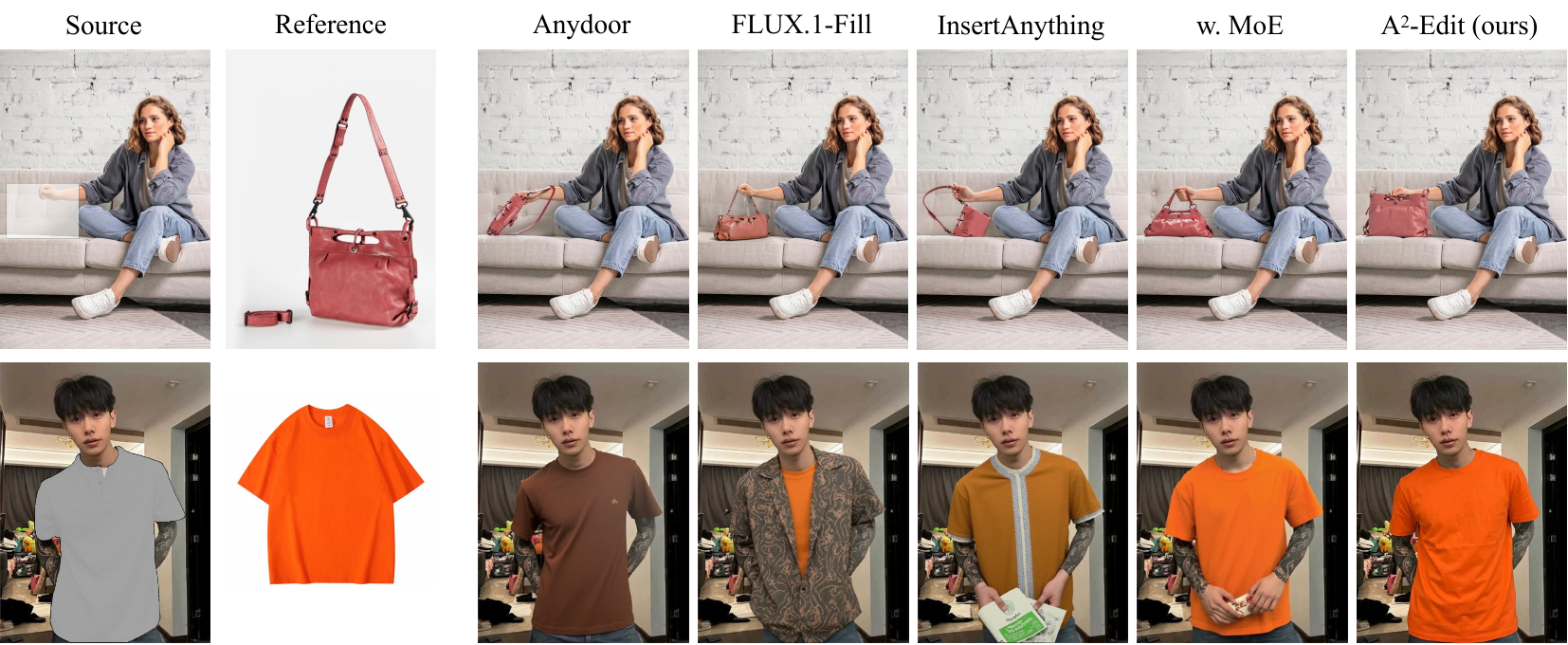}
  % \vspace{-2mm}
  \caption{Garment/Accessory category hold-out experiment results.}
   \label{fig:cross_1}
   \vspace{-2mm}
\end{figure*}

\begin{figure*}[t]
  \centering
  \includegraphics[width=\textwidth]{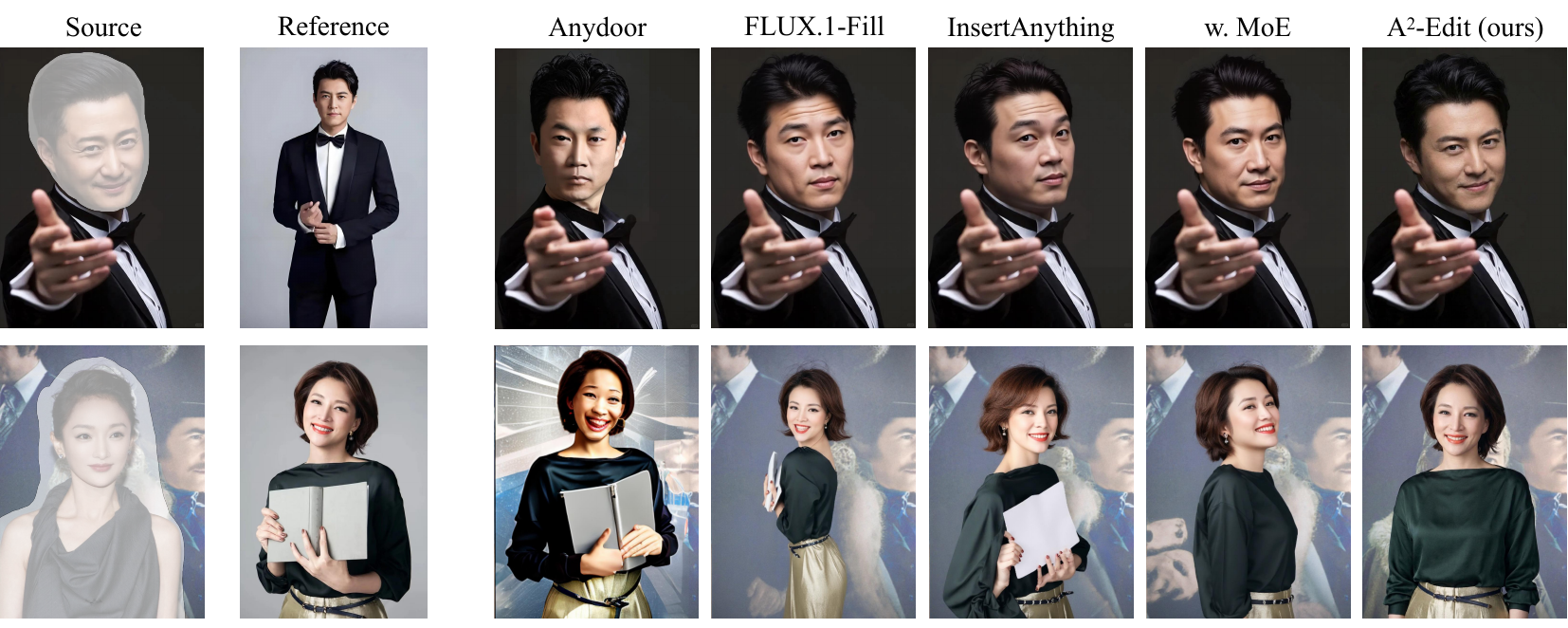}
  % \vspace{-2mm}
  \caption{Portrait category hold-out experiment results.}
   \label{fig:cross_2}
   \vspace{-2mm}
\end{figure*}

\section{Cross-Category Generalization Evaluation}
To further evaluate the cross-domain generalization ability of our framework, we conduct additional experiments under a category hold-out setting. Specifically, we remove all samples belonging to a specific object category during training and evaluate the model on that unseen category during inference. This setup simulates real-world scenarios where a model must generalize to object types that were not observed during training. We design two experimental groups based on the category taxonomy of the UniEdit-500K dataset.

\paragraph{Garment/Accessory Category Hold-out Experiment.}
In the first experiment, we remove all training samples related to garments and accessories, including clothing, shoes, and other wearable items. The model is trained only on the remaining categories. After training, evaluation is conducted exclusively on samples belonging to the garment and accessory categories to assess the model's generalization ability on unseen object types. Figure~\ref{fig:cross_1} presents the qualitative comparison between our method and other baseline models under this setting.

\paragraph{Portrait Category Hold-out Experiment.}
In the second experiment, we remove all portrait-related samples from the training set, including both facial images and full-body human instances, and train the model using only the remaining categories. During evaluation, the model is tested solely on portrait editing tasks. Since portrait editing typically requires strong identity preservation and fine-grained structural consistency, this setting poses a more challenging test of cross-category generalization. Figure~\ref{fig:cross_2} illustrates the qualitative comparison between our method and the baseline approaches under this setting.

In these experiments, we compare our method with several representative image editing approaches, including FLUX.1-Fill, InsertAnything, AnyDoor, and w.moe. To ensure a fair comparison, InsertAnything and AnyDoor are fine-tuned using their official implementations on the same training splits used in our experiments. FLUX.1-Fill is evaluated using its officially released pretrained model for inference, while w.moe denotes a model variant that introduces expert modules only in the feed-forward network (FNN).

The experimental results show that our method consistently outperforms the competing approaches under both hold-out settings. In particular, when the portrait category is completely removed from the training data, most baseline methods suffer from severe identity distortion or structural inconsistencies during testing. In contrast, our model is still able to generate semantically coherent and visually natural editing results. These findings demonstrate that the proposed Mixture-of-Transformers architecture, together with the diverse multi-category training distribution, enables the model to learn transferable editing representations, allowing it to maintain stable performance on previously unseen object categories.

\section{Limitations}
Although A$^2$-Edit demonstrates strong generalization across multi-category editing tasks and under masks of varying precision, it still has several limitations. First, even though MATS enhances the model's robustness to rough masks, it may struggle to accurately infer the editing intent when the provided mask is excessively large. Second, the MoT expert routing mechanism increases memory consumption, requiring more hardware resources during execution. Finally, although UniEdit-500K offers significantly broader category coverage than existing datasets, it still lacks representation for rare object categories and professional fields, which may affect the model's performance in highly specialized scenarios. To address this, we will continue expanding the dataset, with the goal of making it increasingly comprehensive and beneficial to image editing and broader tasks.

\section{Broader Impact}

A$^2$-Edit offers a unified and practical solution for reference-guided image editing, demonstrating strong applicability across a wide range of real-world scenarios. In creative industries such as film production, virtual fashion, and advertising design, the framework enables high-fidelity object replacement and scene editing with few manual effort, substantially improving productivity and lowering the barrier for professional content creation. In e-commerce and virtual try-on applications, A$^2$-Edit can achieve the natural transfer of clothing and products among different subjects, thereby significantly enhancing the user experience. Moreover, its strong tolerance to coarse masks substantially lowers the usage barrier, while its robust generalization across diverse object categories allows it to adapt to a wide range of real-world tasks.

At the same time, similar to other generative models, A$^2$-Edit may pose risks of misuse, such as unauthorized identity manipulation. Therefore, strict ethical guidelines, privacy protection mechanisms, and appropriate regulatory measures are necessary during real-world deployment.

\end{document}